\documentclass[twoside,11pt]{article}

\usepackage[utf8]{inputenc}
\usepackage[T1]{fontenc}
\usepackage{jmlr2e}
\usepackage{booktabs}
\usepackage{bbm}
\usepackage{nicefrac}
\usepackage{microtype}
\usepackage{xcolor}
\usepackage{mathtools}
\mathtoolsset{showonlyrefs}
\usepackage[capitalize,noabbrev]{cleveref}
\usepackage{tikz}
\usepackage{lastpage}
\usepackage{subfig}
\usepackage{etoc}
\usepackage{tcolorbox}

\newtcolorbox{protocolbox}{
    width=0.9\linewidth,
    center,
    colback=yellow!5!white,
    colframe=brown!90!gray,
    boxrule=0.5pt,
    arc=3mm,
    left=2mm,
    right=2mm,
    top=2mm,
    bottom=2mm
}

\renewenvironment{proof}[1][Proof]
  {\par\noindent\textbf{#1}\enspace\ignorespaces}
  {\hfill\BlackBox\par\medskip}

\definecolor{royalblue}{rgb}{0.0, 0.22, 0.66}

\DeclareMathOperator*{\argmax}{arg\,max}

\newtheorem{assumption}{Assumption}

\makeatletter
\@for\thm@x:={theorem,lemma,proposition,remark,corollary}\do{%
  \expandafter\let\csname\thm@x\endcsname\@undefined
  \expandafter\let\csname end\thm@x\endcsname\@undefined
  \expandafter\let\csname c@\thm@x\endcsname\@undefined
  \expandafter\let\csname the\thm@x\endcsname\@undefined
  \expandafter\let\csname p@\thm@x\endcsname\@undefined
}
\makeatother
\newtheorem{theorem}{Theorem}
\newtheorem{lemma}{Lemma}
\newtheorem{proposition}{Proposition}

\newcommand{\dc}{\mathrm{dc}}
\newcommand{\bvartheta}{\boldsymbol{\vartheta}}
\newcommand{\btheta}{\boldsymbol{\theta}}

\jmlrheading{0}{2026}{1-\pageref{LastPage}}{Submitted placeholder}{Published placeholder}{placeholder}{Vahan Arsenyan and Arnak Dalalyan}
\ShortHeadings{Meta-Frieren}{Arsenyan and Dalalyan}
\firstpageno{1}

\begin{document}
\etocdepthtag.toc{main}

\title{Meta-Frieren: Beyond Single Level Online Learning}

\author{\name Vahan Arsenyan \email vahan.arsenyan@ensae.fr \\
       \name Arnak Dalalyan \email arnak.dalalyan@ensae.fr \\
       \addr Department of Statistics\\
       CREST, ENSAE, IP Paris\\
       Rte de Saclay, Palaiseau 91120, France}

\editor{Editor Placeholder}

\maketitle

\begin{abstract}
We study a two-stage online optimization problem: at each round, 
a meta-learner first selects an agent-environment pair, then 
the chosen agent interacts with the environment, and receives a reward.
We introduce Meta-Frieren, a general framework for maximizing cumulative reward
in this setting. Our analysis yields a simple UCB-style criterion for
meta-level selection and recovers prior guarantees as special cases while extending them to broader
settings. We instantiate Meta-Frieren with GP-UCB under Bayesian and frequentist assumptions. We also
derive, to our knowledge, the first high-probability regret bound for Feel-Good
Thompson Sampling, specialize it to finite and parametric function
classes, and integrate it into Meta-Frieren. Experiments support our theoretical 
results and demonstrate that Meta-Frieren can be successfully applied to hyperparameter 
optimization across multiple model classes.

\end{abstract}


\section{Introduction}\label{sec:introduction}

In this paper, we consider an online learning problem in which a
decision-maker allocates observations among several learning agents.
At each round, the decision-maker, referred to as the meta-learner,
selects one of \(K\) agent-environment pairs. The selected agent
observes a context, chooses an action, and receives a noisy reward.
The agents may use different learning algorithms and operate on
different context and action spaces. 
Our goal is to construct a selection rule with small cumulative
regret relative to the agent-environment pair with the largest expected reward
under its optimal policy.

One motivation for this problem comes from hyperparameter
optimization. Suppose that several model classes are available,
each equipped with an optimizer that searches over its
hyperparameters. Given a budget of model evaluations, one has to
decide how many evaluations to allocate to each optimizer. This
decision is complicated by the fact that the performance of an
optimizer depends on the evaluations it has already received.
In particular, a small observed reward can be due either to a poor
choice of model class or to insufficient exploration of its
hyperparameters. An allocation based only on past rewards may therefore assign
too few evaluations to a model class whose hyperparameters
have not yet been sufficiently explored.

The contextual version of this problem has an additional feature:
the meta-learner selects an agent-environment pair before observing its current
context. Thus, it cannot choose a different pair for each observed
context. The value used to compare pairs is consequently the
expectation, over contexts, of the best reward achievable within
each pair. Learning this value requires accounting for the errors
made by the agent, as well as for the variability of the contexts
and the reward noise.

We propose a method, called Meta-Frieren, that constructs an
upper confidence bound for the value of each pair by adding
to its empirical mean reward a bound on the agent's average
regret and concentration terms for context variability and
reward noise.  The resulting
meta-regret bound is expressed in terms of the regret guarantees
available for the individual agents. These guarantees must hold
with high probability, simultaneously over the numbers of visits
to each agent-environment pair, since these numbers are determined by the
meta-learner's previous decisions.

We first establish a general regret bound for this procedure.
The bound accounts for both the selection of suboptimal pairs
and the regret incurred by the agents within the selected pairs.
It holds uniformly over time and allows several pairs to attain
the optimal value. We then study two choices of agents for which
the required local guarantees can be obtained.

The first choice is GP-UCB
\citep{srinivas2009gaussian,krause2011contextual}, which we study
in two settings. In the Bayesian setting, the reward function
\(f_i\) of each pair \(i\) is drawn from a zero-mean Gaussian
process prior with covariance kernel \(k_i\). In the frequentist
setting, \(f_i\) is a deterministic function belonging to the
reproducing kernel Hilbert space (RKHS) associated with \(k_i\).
The kernels \(k_i\) are assumed to be known in both settings.
Our Bayesian result recovers the guarantee of CoCa-BO
\citep{arsenyan2026}, up to constant factors and after accounting
for the difference between the regret definitions.

The second choice is Feel-Good Thompson Sampling
(FGTS) \citep{zhang2022feel}. Here, the available expected-regret
guarantees do not provide the uniform high-probability control
required by the meta-learner. We establish such a guarantee and
give explicit bounds for well-specified finite and
Lipschitz-parametric function classes under the boundedness and
structural assumptions stated below. We also give tuning schemes
that do not require advance knowledge of the horizon and retain
the observations collected in previous rounds. In the linear
case, the resulting regret bound is of order
\(\widetilde O(d\sqrt n)\), where \(d\) is the dimension and \(n\)
is the number of visits. These results allow us to use FGTS agents
within Meta-Frieren and are also of interest for FGTS itself.

We illustrate the results through experiments with GP-UCB and
FGTS agents. These include a comparison with CoCa-BO and an
application to hyperparameter optimization over three model
classes on the MAGIC dataset. In the latter experiment, we also
compare with a greedy allocation rule to assess the effect of
accounting for uncertainty and the agents' learning errors.

The rest of the paper is organized as follows.
After discussing related work, we define the learning protocol
and the regret in \Cref{sec:problem}.
The Meta-Frieren procedure and its general regret guarantee are
presented in \Cref{sec:method}.
\Cref{sec:gp-ucb,sec:fgts} contain the results for GP-UCB and FGTS,
respectively. The experiments are reported in
\Cref{sec:experiments}.

\medskip
\paragraph{Notation}
We generally use capital letters for random variables
and corresponding lowercase letters for their
realizations. Bold letters denote vectors, and
fraktur letters denote context and action spaces.
We write \(\btheta\) for a deterministic parameter
and \(\bvartheta\) for a sampled parameter. Reward
functions \(f_i\), their values \(\mu_i\), and the
gaps \(\Delta_i\) retain the same notation in the
deterministic and Gaussian process settings, although
they are random in the latter. 
All vectors are column vectors, and
\(\mathbf x^{\top}\) denotes transpose.
We write \(\mathbf I\) for the identity matrix,
with dimension determined by context,
\(\mathbb N^+\) for the positive integers, and
\([m]:=\{1,\ldots,m\}\). For vectors, \(\|\cdot\|_p\)
denotes the usual \(\ell_p\) norm; for a function
\(g:\mathcal Z\to\mathbb R\), we write
\(\|g\|_\infty:=\sup_{z\in\mathcal Z}|g(z)|\).
We use \(a\vee b:=\max\{a,b\}\),
\(a\wedge b:=\min\{a,b\}\), and
\(\mathbf1\{A\}\) for the indicator of an event \(A\).
All logarithms are natural.
\section{Related Work}\label{sec:related_work}

Our problem is closely related to contextual causal Bayesian
optimization. In CoCa-BO, introduced by \citet{arsenyan2026},
a policy scope is selected before its context is observed, and
an intervention is then chosen using GP-UCB. The selection of
a scope is based on an average of Gaussian process upper
confidence bounds at previous evaluation points, together with
a correction for context variability. This construction uses
the Gaussian process model at both levels of decision-making.
In Meta-Frieren, the corresponding criterion is computed from
observed rewards, a regret bound for the selected agent, and
concentration radii. The meta-level analysis can therefore be
applied to other agents, provided that suitable regret guarantees
are available. Our Bayesian GP-UCB specialization recovers the
CoCa-BO guarantee up to constant factors, with an adjustment for
the different definitions of regret.

The regret analysis of GP-UCB originates in
\citet{srinivas2009gaussian}, who relate its performance to
the maximum information gain under Gaussian process prior
and RKHS assumptions.
\citet{krause2011contextual} study the contextual setting,
in which an action is chosen after a context has been observed.
These results concern learning within a given environment.
In our setting, they are used to control the regret of an agent
on the rounds when its environment is selected; the allocation
of rounds across environments requires a separate analysis.
For the frequentist version, we use kernel concentration
results of the type established by
\citet{durand2018streaming}.

Our results for posterior sampling build on the FGTS method
of \citet{zhang2022feel}. Its analysis uses decoupling to
bound expected regret for finite and parametric function
classes, including linearly embeddable models. In linear
problems, the resulting rate is
\(\widetilde O(d\sqrt n)\), compared with
\(\widetilde O(d^{3/2}\sqrt n)\) in the classical frequentist
analysis of Thompson sampling by
\citet{abeille2017linear}.
The distinction relevant to the present paper is between
control in expectation and control with high probability.
The number of observations allocated to an FGTS agent is random
and depends on its previous rewards, so the meta-level argument
requires a bound that holds simultaneously over all local
visit counts. We establish such a bound for realized regret.
In the linear specialization, it retains the
\(\widetilde O(d\sqrt n)\) dependence, up to logarithmic factors.

\section{A Two-Stage Learning Problem}
\label{sec:problem}

We consider a framework in which at each round, 
learning involves two connected decisions:
the meta-learner selects an agent-environment pair,
and the selected agent chooses how to act. The first
decision uses the interaction history. Once selected,
the agent observes a context from its environment
and chooses an action. The resulting reward informs
both the meta-learner's future selections and the 
agent's own decisions.

Formally, we consider \(K\) candidate agent-environment
pairs indexed by \(i\in\mathcal I:=[K]\). Each pair has
a context space
\(\mathfrak C_i\subseteq\mathbb R^{d_i^c}\), an action
space \(\mathfrak X_i\subseteq\mathbb R^{d_i^x}\),
and a mean reward function
\(f_i:\mathfrak X_i\times\mathfrak C_i\to\mathbb R\).
After observing a context \(\mathbf c\in\mathfrak C_i\)
and choosing an action \(\mathbf x\in\mathfrak X_i\),
agent \(i\) receives a reward equal to
\(f_i(\mathbf x,\mathbf c)\) plus noise. The objective
is to maximize mean reward by learning both which
pair to select and how to act within each pair.
Crucially, the pair is selected before its context
is observed. 
We study two versions of this framework. In the first,
the mean reward functions \((f_i)_{i\in\mathcal I}\) are
drawn in advance from Gaussian process priors,
independently of the exogenous randomness used to
generate contexts and reward noise. In the second,
the reward functions are deterministic.

The following examples illustrate these two levels
of decision-making.
\vspace*{-3pt}
\begin{itemize}\itemsep=0pt
    \item \textbf{Delivery systems.}
    A logistics company can deploy an autonomous drone,
    a van, or a ground robot. For each delivery, it
    first selects a vehicle. The selected vehicle
    then observes local conditions, such as weather,
    terrain, and traffic, and chooses a route or
    operational action. The reward combines delivery
    time, energy use, and reliability. Each vehicle
    learns how to operate, while the meta-learner
    learns which vehicle performs best overall.

    \item \textbf{Hierarchical hyperparameter
    optimization.}
    A meta-learner first selects a model class.
    An optimizer dedicated to that class then chooses
    a hyperparameter configuration and receives its
    validation performance as a reward. The optimizer
    learns which configurations work well, while the
    meta-learner learns which model class is most
    promising. When no context is present, each context
    space can be a singleton. Our experiment on the
    MAGIC dataset in \Cref{sec:experiments} follows
    this structure, with GBDT, SVM, and \(k\)-NN
    models and their respective hyperparameter spaces.

    \item \textbf{Power-generation technologies.}
    An electricity producer compares several pilot
    plants before expanding deployment. At each round,
    it selects a plant to operate. The plant observes
    current demand, input costs, and equipment
    conditions, then chooses an operating strategy
    for a fixed period. Its reward reflects production,
    cost, and reliability. Each plant learns how to
    operate efficiently, while the meta-learner learns
    which technology is most effective overall.
\end{itemize}

We assume that the relevant functions are measurable
and that the maxima below admit measurable selections.
For each index \(i\), the optimal mean reward at
context \(\mathbf c\) is
\begin{align}
    f_i^\star(\mathbf c)
    :=\max_{\mathbf x\in\mathfrak X_i}
    f_i(\mathbf x,\mathbf c),
    \qquad \mathbf c\in\mathfrak C_i.
\end{align}
An optimal contextual policy for index \(i\) is a
measurable mapping
\(\mathbf x_i^\star:\mathfrak C_i\to\mathfrak X_i\)
that attains this maximum at every context:
\begin{align}
    f_i\big(\mathbf x_i^\star(\mathbf c),\mathbf c\big)
    =f_i^\star(\mathbf c),
    \qquad \text{for all }\mathbf c\in\mathfrak C_i.
\end{align}
Let \(P_i\) denote the unknown distribution over
\(\mathfrak C_i\) from which environment \(i\)
generates contexts, and let \(\mathbf C_i\sim P_i\).
Assuming that \(f_i^\star\) is integrable under
\(P_i\), almost surely in the Gaussian process
setting, define the value of index \(i\) by
\begin{align}
    \mu_i
    :=\mathbb E\big[f_i^\star(\mathbf C_i)\mid f_i\big]
    =\int_{\mathfrak C_i}f_i^\star(\mathbf c)\,
    P_i(\mathrm d\mathbf c).
\end{align}
Thus \(\mu_i\) is the optimal mean reward averaged
over contexts. Conditioning on \(f_i\) makes this
definition applicable to both versions of the
framework: \(\mu_i\) is random in the Gaussian
process setting and deterministic otherwise.

Write \(\boldsymbol f:=(f_1,\ldots,f_K)\).
The random interaction history is
\(H_{t-1}:=((I_s,\mathbf C_s,\mathbf X_s,R_s))_{s=1}^{t-1}\),
with realization
\(h_{t-1}:=((i_s,\mathbf c_s,\mathbf x_s,r_s))_{s=1}^{t-1}\).
Here \(I_s\), \(\mathbf C_s\), \(\mathbf X_s\), and
\(R_s\) denote the selected index, observed context,
chosen action, and reward at round \(s\), respectively.
Starting from an empty history, the interaction
proceeds over a horizon \(T\) that is unknown in
advance.

\begin{protocolbox}
At each round \(t=1,\ldots,T\):
\vspace*{-5pt}
\begin{enumerate}\itemsep=0pt
    \item The meta-learner selects an index \(i_t
    \in [K]\) based on \(h_{t-1}\).

    \item Environment \(i_t\) draws a fresh context
    from \(P_{i_t}\), and the selected agent observes
    its realization \(\mathbf c_t\). Formally,
    \(\mathcal L\big(\mathbf C_t\mid
        \boldsymbol f,H_{t-1},I_t\big)=P_{I_t}\).

    \item The selected agent chooses an action
    \(\mathbf x_t\in\mathfrak X_{i_t}\) based on
    \(h_{t-1}\), \(i_t\), and \(\mathbf c_t\).

    \item The environment reveals a reward \(r_t\),
    a realization of \(R_t\), satisfying
    \begin{align}
        \mathbb E\big[R_t\bigm|
        \boldsymbol f,H_{t-1},I_t,\mathbf C_t,\mathbf X_t\big]
        =f_{I_t}(\mathbf X_t,\mathbf C_t).
    \end{align}%
    \vspace*{-20pt}
    \item The tuple
    \((i_t,\mathbf c_t,\mathbf x_t,r_t)\) is appended
    to the history to obtain \(h_t\).
\end{enumerate}
\end{protocolbox}

The meta-learner's objective is to select indices
attaining the optimal value
\(\mu^\star:=\max_{i\in\mathcal I}\mu_i\), while the
selected agents learn optimal contextual policies.

\paragraph{Regret definition}
Let \(\mathcal I^\star:=\argmax_{i\in\mathcal I}\mu_i\)
denote the set of optimal indices. For each suboptimal
index, define its gap by
\begin{align}\label{eq:gap}
    \Delta_i:=\mu^\star-\mu_i,
    \qquad i\in\mathcal I\setminus\mathcal I^\star,
\end{align}
and write
\(\boldsymbol\Delta:=(\Delta_i)_{i\notin\mathcal I^\star}\).
Let \(N_i(t):=\sum_{s=1}^t\mathbf1\{I_s=i\}\) be the
number of selections of index \(i\) through round \(t\),
with \(N_i(0):=0\). We often refer to this quantity as 
the visit count of \(i\). For \(n\in\mathbb N^+\), 
define the \(n\)th selection time of index \(i\) by
\(\tau_{i,n}
    :=\inf\{t\in\mathbb N^+:N_i(t)=n\}\),
with \(\inf\varnothing:=+\infty\). Variables indexed
by the \(n\)th visit are defined only when
\(\tau_{i,n}<\infty\). 
We use this notation to index random variables and
their realizations at successive selections of 
index \(i\). In particular,
\(\mathbf C_{i,n}:=\mathbf C_{\tau_{i,n}}\),
\(\mathbf X_{i,n}:=\mathbf X_{\tau_{i,n}}\),
\(R_{i,n}:=R_{\tau_{i,n}}\), and
\(\mathbf c_{i,n}:=\mathbf c_{\tau_{i,n}}\).

The within-index regret after \(n\) visits to index
\(i\) is
\begin{align}\label{eq:within_index_regret}
    \mathcal R_{i,n}
    :=\sum_{j=1}^n
    \big(f_i^\star(\mathbf C_{i,j})
    -f_i(\mathbf X_{i,j},\mathbf C_{i,j})\big),
    \qquad \mathcal R_{i,0}:=0.
\end{align}
We define the total cumulative regret by
\begin{align}\label{eq:our_regret}
    \mathcal R(T)
    :=\sum_{i\notin\mathcal I^\star}\Delta_iN_i(T)
    +\sum_{i\in\mathcal I}\mathcal R_{i,N_i(T)}.
\end{align}
The first term measures the loss from selecting
suboptimal indices. The second measures the loss
from choosing suboptimal actions at the contexts
encountered within each selected environment.
Both terms are nonnegative, and \(\mathcal R(T)=0\)
precisely when every selected index is optimal and
every chosen action is optimal for its observed
context. This definition uses mean rewards rather
than observed rewards, so it contains no direct
reward noise term. It remains random through the
contexts, the learner's decisions, and, in the
Gaussian process setting, the reward functions.

Our definition differs from the regret considered
by \citet{arsenyan2026}, which, in our notation, is
\begin{align}\label{eq:regret}
    \widehat{\mathcal R}(T)
    :=\sum_{t=1}^T
    \Big(\mu^\star-\mathbb E\big[
    f_{I_t}(\mathbf X_t,\mathbf C_t)
    \bigm|\boldsymbol f,H_{t-1}\big]\Big).
\end{align}
One can easily check that if \(I_t\) is 
entirely determined by \(H_{t-1}\), we have
\begin{align}\label{eq:weak_regret_open}
    \widehat{\mathcal R}(T)
    &=\sum_{i\notin\mathcal I^\star}\Delta_i
    N_i(T) + \sum_{t=1}^T\mathbb E\big[
    f_{I_t}^\star(\mathbf C_t)
    -f_{I_t}(\mathbf X_t,\mathbf C_t)
    \bigm|\boldsymbol f,H_{t-1}\big].
\end{align}
Thus, the two definitions of regret have the same 
index selection component. Their action selection 
components differ: \(\mathcal R(T)\) evaluates the 
loss at the realized contexts and actions, whereas
\(\widehat{\mathcal R}(T)\) averages this loss
conditionally on the reward functions and preceding
history, including any randomness in action selection.
For a fixed horizon \(T\), assuming integrability,
the tower property yields \( \mathbb E[\,\mathcal 
R(T)\mid\boldsymbol f\,] = \mathbb E[\,\widehat{
\mathcal R}(T)\mid\boldsymbol f\,]\). 
This equality does not imply a sample-path ordering
between the two regrets. In Appendix~\ref{sec:reg_conv},
we establish that our upper bounds also hold for
\(\widehat{\mathcal R}(T)\) up to multiplicative constants under the assumptions
considered by \citet{arsenyan2026}.

\paragraph{Comparison with contextual index selection}
The order of decisions distinguishes our framework
from settings in which contextual information is
available before index selection. Here, the
meta-learner selects an index using historical
information; only then does the selected agent
observe its context and choose an action. The
appropriate oracle benchmark is therefore
\begin{align}
    \mu^\star
    =\max_{i\in\mathcal I}
    \mathbb E\big[f_i^\star(\mathbf C_i)
    \bigm|\boldsymbol f\big].
\end{align}
By contrast, suppose that a jointly distributed
collection of contexts \((\mathbf C_i)_{i\in\mathcal I}\),
independent of \(\boldsymbol f\) and with marginals
\(P_i\), were observed before index selection.
An oracle could then adapt both its index and its
action to the current contexts, attaining
\begin{align}
    \mathbb E\big[\max_{i\in\mathcal I}
    f_i^\star(\mathbf C_i)\bigm|\boldsymbol f\big]
    \geqslant\mu^\star.
\end{align}
The inequality can be strict when different indices
are preferable under different context realizations.
This alternative benchmark corresponds to a richer
information structure and requires a joint context
model, neither of which is assumed in our framework.

\section{Proposed Method and Regret Guarantee}
\label{sec:method}

For each index \(i\), define the empirical mean reward
after \(n\) visits by
\begin{align}
    \widehat\mu_{i,n}
    :=\frac1n\sum_{j=1}^n R_{i,j}.
\end{align}
This quantity reflects both the quality of the
environment and the agent's progress in learning
how to act within it. After visiting each index
once, a naive meta-learner might select at round
\(t\) the index with the largest empirical mean
\(\widehat\mu_{i,N_i(t-1)}\). However, an agent with
a large optimal value may initially appear
unpromising because it has not yet learned an
effective policy. Selecting indices solely by
their empirical rewards may therefore deprive
promising agents of the visits needed to improve.

Meta-Frieren\footnote{The name ``Meta-Frieren''
is a reference to the manga and anime
\emph{Frieren: Beyond Journey's End}. The prefix
``meta'' reflects the hierarchical structure:
the meta-learner learns which agent to select,
while each selected agent learns within its
environment.}
accounts for this learning effect by adding a
within-index regret correction to the empirical
mean, together with confidence radii for context
variability and reward noise. The resulting
upper confidence bound, denoted by
\(\mathrm{UCB}_{i,n}\), provides an optimistic 
estimate of \(\mu_i\) after \(n\) visits. The 
algorithm selects the index with the largest
upper confidence bound, using a fixed 
deterministic tie-breaking rule. We set
\(\mathrm{UCB}_{i,0}:=+\infty\); the bounds for
\(n\geqslant1\) used in the high-level description 
of the algorithm presented below are assumed to be 
finite. Thus every index is visited once before any
index is revisited.

\begin{protocolbox}
\textbf{Meta-Frieren.}\\
Fix a confidence parameter \(\delta\in(0,1)\) and
initialize \(N_i(0)=0\) for every \(i\in\mathcal I\).

At each round \(t=1,2,\ldots\):
\vspace*{-5pt}
\begin{enumerate}\itemsep=0pt
    \item Select the agent corresponding to the index 
    \begin{align}
        I_t\in\argmax_{i\in\mathcal I}
        \mathrm{UCB}_{i,N_i(t-1)}.
    \end{align}%
    \vspace*{-15pt}
    \item The selected agent observes its context,
    chooses an action, and receives a reward
    according to the protocol in \Cref{sec:problem}.

    \item Update the selected agent, its visit count,
    and its empirical mean reward.
\end{enumerate}
\end{protocolbox}

The confidence bounds constructed below make this
procedure independent of the horizon. Its performance
depends on both the choice of these bounds and the
agents' action selection rules. We next state
conditions on the agents' learning performance,
context variability, and reward noise that guide
the construction of \(\mathrm{UCB}_{i,n}\) and yield
a general regret bound. Subsequent sections
specialize this bound to concrete settings in
which regret grows sublinearly with the horizon \(T\).

\subsection{Concentration assumptions used in the regret upper bound}

The definition of within-index regret gives the
decomposition
\begin{align}\label{eq:emp_mean_decomp}
    \widehat\mu_{i,n}
    &=\mu_i - \underbrace{\frac{\mathcal R_{i,n}}n
    \vphantom{\sum_{j=1}^n}}_{
    \mathrm{ALL}}
    +\Big(\underbrace{\frac1n\sum_{j=1}^n
    f_i^\star(\mathbf C_{i,j})-\mu_i}_{\mathrm{OCR}}\Big)
    +\underbrace{\frac1n\sum_{j=1}^n
    \big(R_{i,j}-f_i(\mathbf X_{i,j},\mathbf 
    C_{i,j})\big)}_{\mathrm{RwN}}.
\end{align}
Thus, estimating \(\mu_i\) from observed rewards
requires controlling three quantities: the agent's
learning loss (ALL), fluctuations in the optimal 
contextual reward (OCR), and reward noise (RwN). 
The following assumptions provide these controls 
uniformly over visit counts. The bounds are required 
under the interaction protocol of \Cref{sec:problem}, 
including adaptive selection of the agents.
\begingroup
\renewcommand{\theassumption}{(MF)}
\begin{assumption}\label{assump:regret_terms}
    For each \(i\in\mathcal I\), there exist
    deterministic functions \(B_i\),
    \(\mathrm{rad}_i^{\mathrm{ctx}}\), and
    \(\mathrm{rad}_i^{\mathrm{noise}}\) from
    \(\mathbb N^+\times(0,1)\) to
    \(\mathbb R_{\geqslant0}\) such that, for every
    \(\delta\in(0,1)\), each of the following bounds
    holds with probability at least \(1-\delta\),
    simultaneously over all visit counts:
    \begin{align}
        &&\mathcal R_{i,n}&\leqslant B_i(n,\delta),
        \qquad &&\forall n\in\mathbb N^+,&&\qquad
        \qquad&& \tag{ALL}\label{eq:ALL}\\
        &&\Big|\frac1n\sum_{j=1}^n
        f_i^\star(\mathbf C_{i,j})-\mu_i\Big|
        &\leqslant
        \mathrm{rad}_i^{\mathrm{ctx}}(n,\delta),
        \qquad &&\forall n\in\mathbb N^+,&&~&&
        \tag{OCR}\label{eq:OCR}\\
        &&\Big|\frac1n\sum_{j=1}^n
        \big(R_{i,j}
        -f_i(\mathbf X_{i,j},\mathbf C_{i,j})\big)
        \Big| &\leqslant
        \mathrm{rad}_i^{\mathrm{noise}}(n,\delta),
        \qquad &&\forall n\in\mathbb N^+.&&~&&
        \tag{RwN}\label{eq:RwN}
    \end{align}
\end{assumption}
\addtocounter{assumption}{-1}
\endgroup
For an agent selected only finitely often, these
inequalities are required at the visits actually
made, that is, for \(n\leqslant N_i(\infty)\). 
The functions \(B_i\), \(\mathrm{rad}_i^{
\mathrm{ctx}}\), and \(\mathrm{rad}_i^{\mathrm{
noise}}\) are assumed known to the meta-learner. 
If the horizon \(T\) is known, it suffices to 
require simultaneous validity for visit counts 
up to \(T\).

These assumptions separate the meta-level analysis
from the choice of learning algorithm within each
pair. Within-index regret bounds are available for
GP-UCB under suitable Gaussian process and RKHS
assumptions
\citep{srinivas2009gaussian,krause2011contextual,
iwazaki2025improved}. In \Cref{sec:fgts}, we establish
analogous guarantees for FGTS under deterministic
reward functions and suitable assumptions on the
candidate model class. Context concentration in the
Gaussian process setting is considered in
\citet[Lemma~3]{arsenyan2026}. The noise condition
holds, for example, under a uniform conditional
sub-Gaussian bound or uniformly bounded martingale
differences, with suitable anytime confidence radii.

\subsection{The Meta-Frieren selection rule}

We now use the bounds in \Cref{assump:regret_terms}
to construct an upper confidence bound on each
index's value \(\mu_i\), accounting for both the
agent's learning loss and the variability in
its observed rewards. Fix \(\delta\in(0,1)\) and 
define the abbreviations
\begin{align}
    B_i(n)&:= \max_{1\leqslant s\leqslant n} 
    B_i\big(s,\delta/(3K)\big),\quad
    \mathrm{rad}_i(n)
    :=\mathrm{rad}_i^{\mathrm{ctx}}
    \big(n,\delta/(3K)\big)
    +\mathrm{rad}_i^{\mathrm{noise}}
    \big(n,\delta/(3K)\big).
\end{align}
Applying the three parts \eqref{eq:ALL}, \eqref{eq:OCR}
and \eqref{eq:RwN} of \Cref{assump:regret_terms} with 
confidence parameter \(\delta/(3K)\) and taking a 
union bound over indices gives a common event of 
probability at least \(1-\delta\). On this event, 
the decomposition \eqref{eq:emp_mean_decomp} and the 
nonnegativity of \(\mathcal R_{i,n}\) imply
\begin{align}\label{ineq:mean_ineq}
    \mu_i-\frac{B_i(n)}n-\mathrm{rad}_i(n)
    \leqslant\widehat\mu_{i,n}
    \leqslant\mu_i+\mathrm{rad}_i(n),
    \qquad \forall i\in[K],\ \forall n\in\mathbb 
    N^+.
\end{align}
This motivates the meta-level upper confidence bound
\begin{align}\label{eq:ucb_def}
    \mathrm{UCB}_{i,n}
    :=\widehat\mu_{i,n}+\frac{B_i(n)}n
    +\mathrm{rad}_i(n),
    \qquad n\geqslant1.
\end{align}
The correction \(B_i(n)/n\) accounts for the agent's
learning loss, while \(\mathrm{rad}_i(n)\) accounts
for context and reward fluctuations. In particular,
on the high-probability event \eqref{ineq:mean_ineq},
\begin{align}\label{eq:meta_ucb_sandwich}
    \mu_i\leqslant\mathrm{UCB}_{i,n}
    \leqslant\mu_i+\frac{B_i(n)}n
    +2\,\mathrm{rad}_i(n)\qquad \forall i\in[K],\ 
    \forall n\in\mathbb N^+.
\end{align}
The upper bound in \eqref{eq:meta_ucb_sandwich}
shows how Meta-Frieren limits visits to suboptimal
indices. Once \(B_i(n)/n+2\,\mathrm{rad}_i(n) < 
\Delta_i\), the index's upper confidence bound 
falls below the value of an optimal index. 
For the following result,  set \(B_i(0):=0\).

\begin{theorem}\label{theor:meta_regret}
    Suppose that \Cref{assump:regret_terms}
    holds. For Meta-Frieren, the following 
    statements hold with probability at least 
    \( 1 - \delta\), simultaneously for
    all \(T\in\mathbb N^+\). 
    For every \(i\notin\mathcal I^\star\),
    \begin{align}\label{eq:upper_bound_action}
        N_i(T)\leqslant n_i(T)
        :=\min\Big\{n\in[T]:
        \frac{B_i(n)}n+2\,\mathrm{rad}_i(n)
        <\Delta_i\Big\},
    \end{align}
    with \(n_i(T):=T\) if the set is empty.
    Moreover,
    \begin{align}\label{eq:regret_meta_bound}
        \mathcal R(T)
        &\leqslant
        \sum_{i\notin\mathcal I^\star}
        n_i(T)\Delta_i
        +\sum_{i\notin\mathcal I^\star}
        B_i\big(n_i(T)\big) + \max_{\|\mathbf u\|_1\leqslant T}
        \sum_{i\in\mathcal I^\star}B_i(u_i),
    \end{align}
    where the last maximum is over all 
    \(\mathbf u\in\mathbb N^{\mathcal 
    I^\star}\) whose sum of entries is less than
    or equal to \(T\). 
\end{theorem}

\begin{proof}
Apply each part of \Cref{assump:regret_terms} with
confidence parameter \(\delta/(3K)\), and let
\(\mathcal E\) be the intersection of the resulting
events over all \(i\in\mathcal I\). A union bound
gives \(\mathbb P(\mathcal E)\geqslant1-\delta\).
On this event, simultaneously at all visits made,
\begin{align}
    \mathcal R_{i,n}
    \leqslant B_i\big(n,\delta/(3K)\big)
    \leqslant B_i(n),
\end{align}
and both \eqref{ineq:mean_ineq} and
\eqref{eq:meta_ucb_sandwich} hold. We work on
\(\mathcal E\) throughout the proof.

Fix \(T\in\mathbb N^+\) and a suboptimal index \(i\).
If the set in \eqref{eq:upper_bound_action} is empty,
then \(N_i(T)\leqslant T=n_i(T)\). Otherwise, write
\(n_i:=n_i(T)\) and suppose, for a contradiction,
that \(N_i(T)>n_i\). There is then a round
\(t\leqslant T\) at which
\(N_i(t-1)=n_i\) and \(I_t=i\). By
\eqref{eq:meta_ucb_sandwich} and the definition
of \(n_i\),
\begin{align}
    \mathrm{UCB}_{i,N_i(t-1)}
    \leqslant\mu_i+\frac{B_i(n_i)}{n_i}
    +2\,\mathrm{rad}_i(n_i)
    <\mu_i+\Delta_i=\mu^\star.
\end{align}
For any optimal index \(k\in\mathcal I^\star\),
the lower confidence bound gives
\(\mathrm{UCB}_{k,N_k(t-1)}\geqslant\mu^\star\)
if \(N_k(t-1)\geqslant1\). If \(N_k(t-1)=0\),
its upper confidence bound is infinite.
In either case, selecting \(i\) contradicts the
maximization rule of Meta-Frieren. Therefore,
\(N_i(T)\leqslant n_i(T)\).

The regret decomposition and the within-index
bounds now imply
\begin{align}\label{eq:regret1}
    \mathcal R(T)
    \leqslant
    \sum_{i\notin\mathcal I^\star} N_i(T)\Delta_i
    +\sum_{i\notin\mathcal I^\star}
    B_i\big(N_i(T)\big)
    +\sum_{i\in\mathcal I^\star}
    B_i\big(N_i(T)\big).
\end{align}
By construction, \(B_i(n)\) is nondecreasing in
\(n\), including at zero since \(B_i(0)=0\).
For each suboptimal index, we may therefore replace
\(N_i(T)\) by \(n_i(T)\) in the first two sums.
For the last sum, the vector
\(\mathbf u=(N_i(T))_{i\in\mathcal I^\star}\)
has nonnegative integer entries and satisfies
\(\|\mathbf u\|_1\leqslant T\). It is thus feasible
for the maximum in \eqref{eq:regret_meta_bound},
which proves that inequality.

The argument is pathwise on the same event
\(\mathcal E\) for every \(T\), so both conclusions
hold simultaneously for all horizons.
\end{proof}

The theorem separates regret into the cost of
selecting suboptimal indices, the learning losses
incurred within those indices, and the learning
losses incurred within optimal indices. If
\(B_i(n)/n\to0\) and
\(\mathrm{rad}_i(n)\to0\), each suboptimal index
is selected only finitely many times on a 
high-probability event. Subsequent regret is
therefore driven by learning within the optimal
indices.

The last term allows for multiple optimal indices.
They have the same optimal context-averaged value,
but their learning rates may differ. The maximum
accounts for the worst allocation of visits among
them. When the optimal index is unique, say
\(\mathcal I^\star=\{i^\star\}\), this term reduces
to \(B_{i^\star}(T)\).

To briefly illustrate the usefulness of 
\Cref{theor:meta_regret}, suppose that the 
within-index regret bounds and concentration 
radii satisfy \(B_i(n)=C_1\sqrt n\) and
\(\mathrm{rad}_i(n)=\frac{C_2}{\sqrt n}\), 
for some constants \(C_1,C_2>0\), uniformly over
\(i\in\mathcal I\). Such bounds, possibly up to
logarithmic factors, arise in many standard online
learning settings, including those considered in
the following sections. Under these conditions,
\Cref{theor:meta_regret} yields the regret bound
\begin{align}\label{eq:illustrative_regret}
    \mathcal R(T) \leqslant 4(C_1 + C_2) \sqrt{KT}
    + \|\boldsymbol\Delta\|_1.
\end{align}
Thus, the additional task of learning which agent
to select does not deteriorate the (sublinear in 
\(T\)) \(\sqrt T\)-dependence of the regret bound, 
although it introduces a dependence on the number 
of candidate agents.

To establish this bound, consider a suboptimal
index \(i\notin\mathcal I^\star\). By the definition
of \(n_i(T)\), either \(n_i(T)=1\), or
\begin{align}
    \frac{B_i(n_i(T)-1)}{n_i(T)-1} 
    + \mathrm{rad}_i(n_i(T)-1)\geqslant 
    \Delta_i,
\end{align}
which is equivalent to 
\({C_1+2C_2} \geqslant \Delta_i \sqrt{n_i(T)-1}\). 
Consequently,
\[
    N_i(T)\leqslant n_i(T)
    \leqslant 1+
    \frac{(C_1+2C_2)^2}{\Delta_i^2}
    .
\]
Taking into account this inequality 
and the fact that \(\sum_i N_i(T)\leqslant T\), for every \(b>0\), we get
\begin{align}
    \sum_{i\notin\mathcal I^\star}
    N_i(T) \Delta_i
    &\leqslant \sum_{i:\Delta_i\leqslant b }
    N_i(T) b + 
    \sum_{i:\Delta_i>b}
    \frac{(C_1+2C_2)^2}{\Delta_i}
    +\|\boldsymbol\Delta\|_1
    \leqslant
    bT + \frac{K(C_1+2C_2)^2}b
    +\|\boldsymbol\Delta\|_1.
\end{align}
Optimizing with respect to \(b\), we
get that the regret due to selecting
suboptimal indices satisfies
\begin{align}\label{eq:imr}
    \sum_{i\notin\mathcal I^\star}
    N_i(T) \Delta_i
    \leqslant
    2(C_1+2C_2)\sqrt{KT}
    +\|\boldsymbol\Delta\|_1.
\end{align}
On the other hand, the total within-index learning
loss is bounded, by the Cauchy-Schwarz inequality,
as \(\sum_{i\in\mathcal I}B_i(N_i(T))
    =C_1\sum_{i\in\mathcal I}\sqrt{N_i(T)}
    \leqslant C_1\sqrt{KT}\). 
Combining this inequality 
and \eqref{eq:imr}
with the regret decomposition
\eqref{eq:regret1} proves
\eqref{eq:illustrative_regret}.


\section{Gaussian Process UCB Agents}
\label{sec:gp-ucb}

We now specialize Meta-Frieren to agents that use
Gaussian process upper confidence bounds
(GP-UCB) introduced by~\citet{krause2011contextual}.
The analysis follows the structure of
\Cref{assump:regret_terms}: we specify a within-index
regret bound and confidence radii for context
variability and reward noise, then substitute them
into the general meta-regret guarantee.

We consider two settings: Bayesian and frequentist. 
In the Bayesian setting, the reward functions are 
sampled from Gaussian process priors. In the 
frequentist setting, they are fixed functions 
belonging to a reproducing kernel Hilbert space. 
We also examine the non-contextual case, where sharper
within-index bounds yield improved meta-regret 
guarantees.

For each \(i\in[K]\), let \(k_i\) be a known
covariance kernel on
\(\mathfrak Z_i:=\mathfrak X_i\times\mathfrak C_i\),
normalized so that
\(k_i(\mathbf z,\mathbf z)\leqslant1\) for every
\(\mathbf z\in\mathfrak Z_i\). A central quantity
in the regret analysis of Gaussian process-based
sequential learning algorithms is the maximum
information gain after \(n\) observations
\citep{srinivas2009gaussian,vakili2021information} 
defined by
\begin{align}\label{eq:gamma}
    \gamma_{i,n}
    :=\sup_{\mathbf z_{1:n}\in\mathfrak Z_i^n}
    \frac12\log\det\big(
    \mathbf I_n+\sigma^{-2}
    \mathbf K_i(\mathbf z_{1:n})\big),
\end{align}
where 
\(\mathbf K_i(\mathbf z_{1:n})
:=(k_i(\mathbf z_s,\mathbf z_t))_{s,t=1}^n\)
is the kernel Gram matrix. We set
\(\gamma_{i,0}:=0\). Under a Gaussian process 
prior with independent Gaussian noise, the 
log-determinant expression is the mutual 
information between the function values at 
the evaluation points and their noisy 
observations. Thus \(\gamma_{i,n}\) measures 
the largest information gain obtainable from a 
fixed design of \(n\) evaluations. In the 
frequentist setting, the same expression serves 
as a deterministic measure of kernel complexity.

Throughout this section, we use \(\alpha\in(0,1)\) for a local confidence
parameter and \(\delta\in(0,1)\) for the overall
failure probability of Meta-Frieren. We write
\begin{align}
    \pi_n:={\pi^2n^2}/{3},
    \qquad
    \mathsf C_\sigma :=8/{\log(1+\sigma^{-2})}.
\end{align}

\subsection{Bayesian setting}
\label{sec:by_gp_ucb}

We first consider the Bayesian setting, in which each 
reward function \(f_i\) is drawn from a zero-mean Gaussian 
process with covariance kernel \(k_i\). To derive regret 
guarantees, we impose the following assumptions on the 
kernels, the context distributions, and the observation 
noise. In particular, we require sufficient regularity 
of the Gaussian process sample paths.

\begingroup
\renewcommand{\theassumption}{(BS)}
\begin{assumption}\label{assump:BS}
    For every index \(i=1,\ldots,K\), let \(\mathfrak 
    X_i\subseteq[0,w_i^x]^{d_i^x}\) and \( 
    \mathfrak C_i\subseteq[0,w_i^c]^{d_i^c}\) be
    nonempty, convex, compact sets, and write 
    \(d_i:=d_i^x+d_i^c\) and \(w_i:=w_i^x\vee w_i^c\).
    \begin{enumerate}\itemsep=-2pt
        \item Each \(f_i\) is a Gaussian process 
        on \(\mathfrak X_i\times \mathfrak C_i\), 
        with zero mean and a covariance kernel 
        \(k_i\) bounded by \(1\): 
        \(k_i((\mathbf x,\mathbf c),(\mathbf x,\mathbf c))\leqslant 1\) 
        for every \((\mathbf x,\mathbf c)\in\mathfrak X_i\times 
        \mathfrak C_i\). 

        \item The contexts
        \((\mathbf C_{i,j})_{j\geqslant1}\)
        are iid with distribution \(P_i\),
        and the noise variables
        \((\varepsilon_{i,j})_{j\geqslant1}\)
        are iid \(\mathcal N(0,\sigma^2)\).
        All context and noise sequences are
        mutually independent, both within and
        across indices, and are jointly
        independent of the collection of
        Gaussian processes \((f_i)_{i\in[K]}\).
        At the \(j\)-th visit to index \(i\),
        the reward is
        \(R_{i,j}
            =f_i(\mathbf X_{i,j},\mathbf C_{i,j})
            +\varepsilon_{i,j}\). 

        \item There exist constants
        \(\psi\geqslant1\) and \(\varphi>0\)
        such that, for every \(i\in[K]\)
        and \(\alpha\in(0,1)\), with
        probability at least \(1-\alpha\),
        the sample path \(f_i\) is continuously
        differentiable and
        \begin{align}\label{eq:grad_op_norm_f}
            \|\nabla f_i(\mathbf x,\mathbf c)\|_\infty
            \leqslant
            \varphi\sqrt{\log(d_i\psi/\alpha)},\qquad 
            \forall (\mathbf x,\mathbf c)\in
            \mathfrak X_i\times\mathfrak C_i.
        \end{align}
    \end{enumerate}
\end{assumption}
\addtocounter{assumption}{-1}
\endgroup

Assumptions of this type are standard in Bayesian
optimization. The main regularity requirements
concern the boundedness of the covariance kernels
and the control of the sample-path derivatives in
\eqref{eq:grad_op_norm_f}. A sufficient condition
for the latter is that the covariance kernels are
four times continuously differentiable 
\citep{ghosal2006posterior}.
Importantly, the sample-path regularity condition
plays two roles in our analysis: it controls the
discretization error in GP-UCB and the variability
of the optimal reward across contexts.

At each visit, agent \(i\) selects its action using
the contextual GP-UCB rule. Let
\(m_{i,n-1}(\mathbf x,\mathbf c)\) and
\(s_{i,n-1}(\mathbf x,\mathbf c)\) denote,
respectively, the posterior mean and standard
deviation of \(f_i(\mathbf x,\mathbf c)\), given
the observations collected during the first
\(n-1\) visits to index \(i\).
At its \(n\)-th visit, after observing the
context \(\mathbf C_{i,n}\), agent \(i\)
selects
\begin{align}\label{eq:gp_ucb_action}
    \mathbf X_{i,n}\in
    \argmax_{\mathbf x\in\mathfrak X_i}
    \Big\{
        m_{i,n-1}(\mathbf x,\mathbf C_{i,n})
        +\sqrt{\beta_i(n,\alpha)}
        \,s_{i,n-1}(\mathbf x,\mathbf C_{i,n})
    \Big\}.
\end{align}
The exploration parameter is chosen as
\begin{align}\label{eq:beta_t_gp_ucb}
    \beta_i(n,\alpha)
    :=2\log(2\pi_n/\alpha)
    +2d_i\log_+\Big(
    n^2d_iw_i\varphi
    \sqrt{\log(4d_i\psi/\alpha)}
    \Big),
\end{align}
where \(\log_+(u):=\max\{0,\log u\}\).

We now specify the within-index regret bound and
the two concentration radii required by
\Cref{assump:regret_terms}. For every \(i\in[K]\),
\(n\geqslant1\), and \(\alpha\in(0,1)\), set
\begin{align}\label{eq:b_gpucb_rad}
    \mathrm{rad}_i^{\mathrm{noise}}(n,\alpha)
    :=\sigma\sqrt{
        \frac{2\log(\pi_n/\alpha)}{n}},
    \quad 
    \mathrm{rad}_i^{\mathrm{ctx}}(n,\alpha)
    :=\frac{d_i^cw_i^c\varphi}{\sqrt{2n}}
    \log\frac{2(d_i\psi+\pi_n)}{\alpha},
\end{align}
and 
\begin{align}
    B_i(n,\alpha)
    :=\sqrt{\mathsf C_\sigma n
    \beta_i(n,\alpha)\gamma_{i,n}}+2.
    \label{eq:b_gpucb_B}
\end{align}
We also set \(B_i(0,\alpha):=0\).

The noise radius follows from Gaussian
concentration. The context radius combines
the sample-path regularity condition
\eqref{eq:grad_op_norm_f} with concentration
of the empirical averages of the optimal
contextual rewards. Finally, the within-index
regret bound follows from the analysis of
GP-UCB
\citep{srinivas2009gaussian,krause2011contextual}.
Together, these bounds verify the three
requirements of \Cref{assump:regret_terms}.

Fix \(\delta\in(0,1)\) and set
\(\alpha:=\delta/(3K)\). Each GP-UCB agent
uses the exploration parameter
\(\beta_i(n,\alpha)\), and Meta-Frieren
uses the corresponding within-index regret
bounds and concentration radii.
As in \Cref{sec:method}, we suppress
the confidence argument and write
\begin{align}\label{eq:betaT}
    \mathrm{rad}_i(n)
    &:=\mathrm{rad}_i^{\mathrm{ctx}}(n,\alpha)
    +\mathrm{rad}_i^{\mathrm{noise}}(n,\alpha),
    \qquad
    \rho_i(T)
    :=2\sqrt T\,\mathrm{rad}_i(T).
\end{align}
To state the regret guarantee, define
\begin{align}\label{eq:gammabar}
    \beta(T)
    &:=\max_{i\in[K]}\beta_i(T,\alpha),
    \qquad 
    \gamma(T)
    :=\max_{i\notin\mathcal I^\star}
    \gamma_{i,T},
    \qquad
    \rho(T):=\max_{i\in[K]}\rho_i(T),
\end{align}
where \(\gamma(T):=0\) if every index is
optimal. We also introduce
\begin{align}
    \bar\gamma(T)
    :=\max_{\substack{
    {\mathbf n\in\mathbb N^{\mathcal I^\star}}, \ 
    \|\mathbf n\|_1\leqslant T}}
    \frac1{|\mathcal I^\star|}
    \sum_{i\in\mathcal I^\star}\gamma_{i,n_i}.
\end{align}
The quantities \(\gamma(T)\) and
\(\bar\gamma(T)\) control the information
gain at suboptimal and optimal indices,
respectively.

\begin{theorem}\label{th:bay_gp_ucb}
    Consider Meta-Frieren with the
    meta-level selection rule of
    \Cref{sec:method}, where each agent
    follows the contextual GP-UCB rule
    \eqref{eq:gp_ucb_action} with exploration
    parameter \eqref{eq:beta_t_gp_ucb}.
    Under \Cref{assump:BS}, for every
    \(\delta\in(0,1)\), with probability
    at least \(1-\delta\), simultaneously
    for all \(T\geqslant1\), the cumulative
    regret \(\mathcal R(T)\) defined in~\eqref{eq:our_regret} satisfies
    \begin{align}
        \mathcal R(T)
        &\leqslant
        \sqrt{
            T|\mathcal I^\star|
            \mathsf C_\sigma
            \beta(T)\bar\gamma(T)}
        +
        4\big(
            \mathsf C_\sigma\beta(T)\gamma(T)
            +\rho^2(T)
        \big)
        \sum_{i\notin\mathcal I^\star}
        \frac1{\Delta_i}
        +
        2K\big(2+\sqrt{\beta(1)}\big)
        +\|\boldsymbol\Delta\|_1.
    \end{align}
\end{theorem}

The first term accounts for learning
within optimal indices, whereas the
gap-dependent term controls the cost
of selecting and learning at suboptimal
indices. The remaining terms are
finite-visit corrections.

To illustrate the order of magnitude
of the bound, consider squared-exponential
kernels with fixed hyperparameters.
Writing \(d:=\max_{i\in[K]}d_i\), the
standard information-gain estimate gives
\(\gamma_{i,T}=O((\log(eT))^{d_i+1})\),
see \citet[Theorem~5]{srinivas2009gaussian}.
Moreover, for fixed confidence level,
\(\beta(T)=O(\log(eT))\) and
\(\rho(T)=O(\log(eT))\).
Consequently, \Cref{th:bay_gp_ucb}
implies
\begin{align}\label{eq:bay_gpucb_se_rate}
    \mathcal R(T)
    =\widetilde O\Big(
        \sqrt{T|\mathcal I^\star|}
        +\sum_{i\notin\mathcal I^\star}
        \frac1{\Delta_i}
        +K+\|\boldsymbol\Delta\|_1
    \Big).
\end{align}
Here the notation \(\widetilde O\)
suppresses polylogarithmic factors in
\(T\), with constants depending on the
fixed problem parameters and confidence
level. In particular, the contribution
of optimal indices grows at most as
\(\sqrt T\), up to logarithmic factors,
while the gap-dependent contribution
grows only polylogarithmically with
the horizon.

\subsection{Frequentist setting}
\label{sec:freq_gp_ucb}

We now turn to the frequentist setting, in which
the reward functions are fixed rather than
sampled from Gaussian process priors. More
precisely, we assume that each \(f_i\) belongs
to the RKHS associated with \(k_i\), with
known norm bound \(\|f_i\|_{k_i}\leqslant b\).
We also relax the Gaussian noise assumption,
requiring instead that
\begin{align}
    \mathbb E\big[
        e^{\rho\varepsilon_{i,j}}
        \bigm|
        H_{\tau_{i,j}-1},
        \mathbf C_{i,j},\mathbf X_{i,j}
    \big]
    \leqslant e^{\sigma^2\rho^2/2},
    \qquad \forall\rho\in\mathbb R.
    \label{eq:freq_set:sub_gaussian_noise}
\end{align}
Thus, the observation noise is conditionally
\(\sigma\)-sub-Gaussian.

Each agent continues to select its actions
according to the GP-UCB rule
\eqref{eq:gp_ucb_action}, using the same
kernel-based prediction formulas. In the
present setting, however, these formulas
are viewed as kernel-based estimators,
rather than as the posterior mean and
variance of a correctly specified
Gaussian process model. The exploration
parameter is chosen as
\begin{align}\label{eq:beta_i}
    \beta_i(n,\alpha)
    :=\left(
        b+\sqrt{
            2\gamma_{i,n}
            +2\log(1/\alpha)}
    \right)^2.
\end{align}

We next specify the three bounds required
by \Cref{assump:regret_terms}.
By the reproducing property and the
normalization of the kernel,
\begin{align}\label{eq:freq_function_bound}
    |f_i(\mathbf x,\mathbf c)|
    &\leqslant
    \|f_i\|_{k_i}
    \sqrt{k_i\big(
        (\mathbf x,\mathbf c),
        (\mathbf x,\mathbf c)
    \big)}
    \leqslant b.
\end{align}
In particular, the optimal contextual
rewards \(f_i^\star(\mathbf C_{i,j})\)
take values in \([-b,b]\), so their
empirical averages satisfy a uniform
concentration inequality based on
Hoeffding's inequality. Combined with
the concentration of the sub-Gaussian
observation noise, this leads to
\begin{align}\label{eq:b_gpucb_rad1}
    \mathrm{rad}_i^{\mathrm{noise}}(n,\alpha)
    &:=\sigma\sqrt{
        \frac{2\log(\pi_n/\alpha)}n},
    \qquad
    \mathrm{rad}_i^{\mathrm{ctx}}(n,\alpha)
    :=b\sqrt{
        \frac{2\log(\pi_n/\alpha)}n}.
\end{align}
Moreover, \Cref{prop:cgp-regret} in Appendix~\ref{sec:proof_of_freq_gp_ucb} provides
the within-index regret bound
\begin{align}\label{eq:b_gpucb_B1}
    B_i(n,\alpha)
    :=\sqrt{
        \mathsf C_\sigma n
        \beta_i(n,\alpha)\gamma_{i,n}}.
\end{align}
These choices establish the three
requirements of \Cref{assump:regret_terms}.
Notice that, unlike in the Bayesian
setting, no sample-path regularity
assumption such as
\eqref{eq:grad_op_norm_f} is needed.

Fix \(\delta\in(0,1)\) and set
\(\alpha:=\delta/(3K)\).
As in the Bayesian setting, let
\(\mathrm{rad}_i(n)\) be the sum of
the two concentration radii, and
define \(\beta(T)\), \(\gamma(T)\),
\(\bar\gamma(T)\), and \(\rho(T)\) 
by \eqref{eq:betaT} and \eqref{eq:gammabar}
using the present choices of
\(\beta_i\), \(B_i\), and the radii, 
given by formulas \eqref{eq:beta_i}, 
\eqref{eq:b_gpucb_B1} and 
\eqref{eq:b_gpucb_rad1}, respectively.

\begin{theorem}\label{th:freq_gp_ucb}
    Consider Meta-Frieren with the
    meta-level selection rule of
    \Cref{sec:method}, where each
    agent follows the GP-UCB rule
    \eqref{eq:gp_ucb_action} with
    the exploration parameter
    specified above. Under the
    frequentist assumptions of this
    subsection, for every
    \(\delta\in(0,1)\), with probability
    at least \(1-\delta\),
    simultaneously for all
    \(T\geqslant1\), the cumulative
    regret \(\mathcal R(T)\) defined
    in \eqref{eq:our_regret} satisfies
    \begin{align}
        \mathcal R(T)
        &\leqslant
        \sqrt{
            T|\mathcal I^\star|
            \mathsf C_\sigma
            \beta(T)\bar\gamma(T)}
        +
        4\big(
            \mathsf C_\sigma\beta(T)\gamma(T)
            +\rho^2(T)
        \big)
        \sum_{i\notin\mathcal I^\star}
        \frac1{\Delta_i}+
        2b(K-|\mathcal I^\star|)
        +\|\boldsymbol\Delta\|_1.
    \end{align}
\end{theorem}

The proof is given in
Appendix~\ref{sec:proof_of_freq_gp_ucb}.
As in the Bayesian setting, the first
term accounts for learning within
optimal indices, while the
gap-dependent term controls the
contribution of suboptimal indices.
The main difference is that the
frequentist exploration parameter
depends explicitly on the maximum
information gain, which affects the
resulting rates for particular kernels.

For squared-exponential kernels with fixed
hyperparameters, as mentioned earlier, 
\citet[Theorem~5]{srinivas2009gaussian} shows that 
\(\gamma_{i,T}=O((\log(eT))^{d_i+1})\).
Writing \(d:=\max_{i\in[K]}d_i\), we therefore
have
\(\beta(T)=O((\log(eT))^{d+1})\) and
\(\rho(T)=O(\sqrt{\log(eT)})\), for fixed
confidence level and problem parameters.
Consequently, \Cref{th:freq_gp_ucb} yields
\begin{align}
    \mathcal R(T)
    =\widetilde O\Big(
        \sqrt{T|\mathcal I^\star|}
        +\sum_{i\notin\mathcal I^\star}
        \frac1{\Delta_i}
        +K+\|\boldsymbol\Delta\|_1
    \Big).
\end{align}
Thus, the frequentist and Bayesian settings
lead to the same regret order up to
polylogarithmic factors in \(T\).
The frequentist bound has larger logarithmic
factors because its exploration parameter
depends on the maximum information gain.

\section{Feel-Good Thompson Sampling}\label{sec:fgts}

We now specialize Meta-Frieren to agents that use Feel-Good
Thompson Sampling (FGTS) introduced in~\citet{zhang2022feel}. 
FGTS balances
predictive accuracy and optimism: its sampling distribution
favours candidate models that fit past observations while
predicting large optimal rewards. To integrate these agents
into Meta-Frieren, we establish high-probability bounds on
their realized regret that hold uniformly over time.

We first specify the sampling rule and its assumptions.
We then establish high-probability regret bounds for
well-specified models with finite and continuous parameter 
spaces, and provide explicit tuning schemes that do not 
require knowledge of the horizon. These results yield the
Meta-Frieren guarantees presented at the end of the section.

\subsection{Model and sampling rule}\label{sec:fgts_setup}
For each \(i\in[K]\), let \(\Theta_i\) be a parameter space,
\(p_{i,0}\) a prior probability measure on \(\Theta_i\), and
\(\{f_{i,\btheta}:\btheta\in\Theta_i\}\) a family of candidate
reward functions. All functions and kernels below are assumed
measurable, and the maxima are assumed to admit measurable
selections. Fix a measurable tie-breaking rule and define
\begin{align}\label{eq:fgts_greedy_action}
    \mathbf x_i(\btheta,\mathbf c)
    &\in\argmax_{\mathbf x\in\mathfrak X_i}
    f_{i,\btheta}(\mathbf x,\mathbf c),
    \qquad f_{i,\btheta}(\mathbf c)
    :=f_{i,\btheta}\big(\mathbf x_i(\btheta,\mathbf c),\mathbf c\big).
\end{align}
Thus, \(f_{i,\btheta}(\mathbf c)\) is the optimal reward
predicted by candidate \(\btheta\) at context \(\mathbf c\).

At its \(j\)th visit, agent \(i\) observes \(\mathbf C_{i,j}\),
draws \(\bvartheta_{i,j}\) from the sampling distribution
\(p_{i,j}\) defined below, and plays
\(\mathbf X_{i,j}=\mathbf x_i(\bvartheta_{i,j},\mathbf C_{i,j})\).
It then receives a noisy reward \(R_{i,j}\) with conditional
mean \(f_i(\mathbf X_{i,j},\mathbf C_{i,j})\). The observed
context, action, and reward are recorded in the interaction
history and used to update the sampling distribution for
subsequent visits to this agent-environment pair.

To specify the policy, it remains to define the distribution
\(p_{i,j}\). To this end, we choose a fixed \(\eta_i>0\)
and a deterministic, positive, nonincreasing sequence
\((\lambda_{i,j})_{j\geqslant1}\), and define the ``energy''
\begin{align}\label{eq:fgts_neg_log_likelihood}
    L_{i,j}(\btheta,\mathbf c,\mathbf x,r)
    :=\eta_i\big(f_{i,\btheta}(\mathbf x,\mathbf c)-r\big)^2
    -\lambda_{i,j}f_{i,\btheta}(\mathbf c).
\end{align}
FGTS uses the Gibbs posterior
\begin{align}\label{eq:fgts_posterior}
    p_{i,j}(\mathrm d\btheta)
    \propto\exp\Big(-\sum_{s=1}^{j-1}
    L_{i,s}(\btheta,\mathbf C_{i,s},\mathbf X_{i,s},R_{i,s})\Big)
    p_{i,0}(\mathrm d\btheta),
\end{align}
where we suppress the dependence on the interaction history.
Each past observation retains the coefficient \(\lambda_{i,s}\)
assigned to its visit.
The squared-loss term favours candidates that predict past
rewards accurately, while the feel-good term favours candidates
with large predicted optimal rewards. The resulting Gibbs
posterior need not coincide with a Bayesian posterior for
the reward noise.

We analyse this sampling rule under the following 
assumptions on the noise, the candidate predictions, 
and the expected reward. We first assume that the 
reward noise is conditionally sub-Gaussian with 
variance proxy one.
\begingroup
\renewcommand{\theassumption}{(FGTS-N)}
\begin{assumption}\label{ass:FGTS-N}
    Specifically, for every \(i\in[K]\)
    and \(j\geqslant1\),
    \begin{align}\label{eq:fgts_noise}
        R_{i,j}=f_i(\mathbf X_{i,j},\mathbf C_{i,j}) 
        +\varepsilon_{i,j}, \qquad
        \mathbb E\big[e^{\rho\varepsilon_{i,j}}\bigm|
        H_{\tau_{i,j}-1},\mathbf C_{i,j},\mathbf X_{i,j}
        \big] \leqslant e^{\rho^2/2},
        \quad \forall\rho\in\mathbb R.
    \end{align}    
\end{assumption}
\addtocounter{assumption}{-1}
\endgroup
This condition implies that the noise has conditional mean
zero. It accommodates adaptive action selection and does
not require independence of the noise variables.
\begingroup
\renewcommand{\theassumption}{(FGTS-F)}
\begin{assumption}\label{ass:FGTS-F}
    The true and candidate reward functions are 
    deterministic and satisfy the following conditions 
    for every \(i\in[K]\). 
    \begin{enumerate}
        \item  For every 
        \(\btheta\in\Theta_i\) and \((\mathbf x,\mathbf
        c)\in\mathfrak X_i\times \mathfrak C_i\),
        \begin{align}\label{eq:fgts_bounded_rewards}
            0\leqslant f_i(\mathbf x,\mathbf c)\leqslant
            1, \qquad 0\leqslant f_{i,\btheta}
            (\mathbf x,\mathbf c) \leqslant 1.
        \end{align}
        \item There are an integer \(m_i\geqslant1\), 
        measurable functions \(u_{i,1},\ldots,u_{i,m_i}:
        \Theta_i\times\mathfrak C_i\to\mathbb R\), and 
        measurable functions \(\phi_{i,1},\ldots,
        \phi_{i,m_i}: \mathfrak X_i\times\mathfrak C_i\to 
        \mathbb R\) such that
        \begin{align}
            f_{i,\btheta}(\mathbf x,\mathbf c)-f_i(
            \mathbf x,\mathbf c) = \sum_{\ell=1}^{m_i}
            u_{i,\ell}(\btheta,\mathbf c)
            \phi_{i,\ell}(\mathbf x,\mathbf c), \qquad
            (\btheta, \mathbf x,\mathbf c)\in \Theta_i
            \times \mathfrak X_i\times \mathfrak C_i.
        \end{align}
        For each fixed \(\mathbf c\), the functions 
        \(u_{i,\ell}(\cdot,\mathbf c)\) are bounded.
    \end{enumerate}   
\end{assumption}
\addtocounter{assumption}{-1}
\endgroup
Under the bounds required by the first condition above, 
the clipping of the feel-good term and the posterior support
restrictions used in the analysis of \citet{zhang2022feel} 
are unnecessary. Bounded candidate predictions also ensure 
that the normalizing constant in~\eqref{eq:fgts_posterior} 
is finite and strictly positive at every finite visit. 
Boundedness of the candidate predictions can be enforced
by restricting the parameter space to candidates taking
values in \([0,1]\), provided that the restricted
family retains a representation of the true reward
function. This restriction preserves the linear
representation in the second condition. The prior is
then chosen on the restricted parameter space.

The final assumption specifies the tuning conditions
for the energy in \eqref{eq:fgts_neg_log_likelihood}.
\begingroup
\renewcommand{\theassumption}{(FGTS-P)}
\begin{assumption}\label{ass:FGTS-P}
    For every \(i\in [K]\),
    \vspace{-5pt}
    \begin{enumerate}\itemsep=0pt
        \item \(\eta_i\) is deterministic, positive, and
        satisfies \(4\eta_i \leqslant 1\).
        \item The sequence \( ( \lambda_{i,j} )_{j\geqslant
        1}\) is deterministic, positive, and nonincreasing. 
    \end{enumerate}
\end{assumption}
\addtocounter{assumption}{-1}
\endgroup
For the bounds below, define
\begin{align}
    \bar\kappa_n :=\max_{i\in[K]}\frac1n\sum_{j=1}^n
    \big(\lambda_{i,j}/\lambda_{i,n}\big)^2,\qquad 
    \bar\lambda_{i,n} := \frac1n\sum_{j=1}^n\lambda_{i,j}.
\end{align}

\subsection{High-probability regret bounds in 
well-specified discrete case}

In this section, we assume that the true noiseless
reward function \(f_i\) belongs to the parametric
class \(\{f_{i,\btheta}:\btheta\in\Theta_i\}\). We 
refer to this case as a well-specified model. We 
consider here the case of discrete  
parameter spaces and treat the continuous parameter
spaces in the next section. 

\begin{theorem}\label{th:fgts_discrete}
    Suppose that Assumptions~\ref{ass:FGTS-F},~\ref{ass:FGTS-N} and~\ref{ass:FGTS-P} hold. Fix 
    \(i\in[K]\) and \(\delta\in(0, 1/2)\).  
    Suppose that \(\Theta_i\) is countable and that 
    \( q_i^\star:=p_{i,0}\big(\{\btheta : 
    f_{i,\btheta}\equiv f_i\} \big)>0\). With probability 
    at least \(1-\delta\), the following bound holds
    simultaneously for all \(n\geqslant1\):
    \begin{align}\label{eq:fgts_regret_b}
        \mathcal R_{i,n}
        &\leqslant \frac{m_i\bar\kappa_n}{\eta_i} 
        \lambda_{i,n} n + \frac{\log(2/\delta) 
        + \log(1/q_i^\star)}{\lambda_{i,n}} +
        \sqrt{n \log({2n}/{\delta})} .
    \end{align}
\end{theorem}

The proof is deferred to \Cref{sec:fgts_discr}. 
We can use this bound to choose the FGTS parameters.
Suppose that, for every \(i\in[K]\), the parameter space
\(\Theta_i\) is finite and \(p_{i,0}\) is uniform on
\(\Theta_i\). Assuming \(m_i\) is known, choose
\begin{align}\label{eq:choice}
    \eta_i=\frac14,\qquad
    \lambda_{i,n}
    =\sqrt{\frac{
        \smash{\log(2|\Theta_i|/\delta)}\vphantom{\log a}
    }{4m_i n}}.
\end{align}
For this choice, \(\bar\kappa_n=\sum_{j=1}^n j^{-1}
\leqslant1+\log n\). Substituting \eqref{eq:choice}
into \eqref{eq:fgts_regret_b} and simplifying shows that,
for each fixed \(i\in[K]\), with probability at least
\(1-\delta\), simultaneously for all \(n\geqslant1\),
\begin{align}\label{eq:regr4}
    \mathcal R_{i,n} \leqslant 2.5\log(6n)
    \sqrt{nm_i\log(2|\Theta_i|/\delta)}.
\end{align}
The factor \(\log n\) in~\eqref{eq:regr4} results from
the chosen horizon-independent tuning and the accumulation
of past coefficients in our bound. If the horizon \(N\) were known, we could instead choose
\begin{align}
    \lambda_{i,n}
    =
    \sqrt{\frac{\smash{\log(4|\Theta_i|/\delta)}
    \vphantom{\log a}}{4m_i N}},
    \qquad n\leqslant N.
\end{align}
Then \(\bar\kappa_N=1\), eliminating the harmonic-number
factor from the bound at \(N\). The concentration term
still contains \(\sqrt{\log(2N/\delta)}\).
We favour a simple bound here over optimized logarithmic
factors. Replacing \(m_i n\) in~\eqref{eq:choice} by
\(m_i n\log(2n)\) improves the asymptotic factor
\(\log n\) in~\eqref{eq:regr4} to \(\sqrt{\log n}\, 
\log\log n\). 

An alternative way to remove dependence on the horizon
is the doubling trick: each agent runs successive
epochs of lengths \(1,2,4,\ldots\) on its local clock,
restarting FGTS with parameters tuned to the length
of each epoch. Geometric restarts preserve
square-root regret rates up to constant factors,
apart from logarithmic terms already present in the
bounds; see \citet[Exercise~6.6]{lattimore2020bandit}
and \citet{besson2018doubling}. Starting from
high-probability bounds, a guarantee valid
simultaneously over all epochs can be obtained by
allocating a summable sequence of failure
probabilities across epochs. This approach allows
one to reuse a fixed-horizon analysis and can yield
smaller logarithmic factors than our simple tuning.
However, restarting the posterior discards the
information accumulated in previous epochs and
may lead to repeated exploration. Our
round-dependent tuning retains all past observations
and provides a guarantee uniform over visit counts
without restarting the algorithm.

The upper bound in \eqref{eq:regr4} 
has leading dependence \(\sqrt{m_i n\log|\Theta_i|}\), 
apart from confidence- and horizon-dependent 
logarithmic factors. When \(\mathfrak X_i\) contains 
\(M_i\) actions, we may take \(m_i=M_i\). In this 
specialization, our bound matches, up to logarithmic 
factors, the minimax regret rate for general 
realizable contextual bandits with finite action 
and function classes; see \citet{foster2020beyond}. 
This is a worst-case optimality statement over such 
classes, rather than an optimality statement for each 
individual class.

\subsection{High-probability regret bounds in 
well-specified continuous case}\label{sec:fgts_param_fnc}

We now consider continuous parameter spaces and candidate
reward functions that depend Lipschitz-continuously on
the parameter. For a uniform prior on the unit Euclidean
ball, we obtain the following result proved in 
\Cref{sec:fgts_cont}.

\begin{theorem}\label{th:fgts_cont}
    Suppose that Assumptions~\ref{ass:FGTS-N},~\ref{ass:FGTS-F}, and~\ref{ass:FGTS-P} hold. 
    Fix \(i\in[K]\) and \(\delta\in(0,1/2)\). Suppose that
    \(\Theta_i=\mathbb B_{1}(0)\subseteq \mathbb R^{ 
    d_i^\theta}\), and that \(p_{i,0}\) is uniform on 
    \(\Theta_i\). Assume that \(f_i\in\{f_{i,\btheta}:
    \btheta\in\Theta_i\}\) and that
    \begin{align}
        \|f_{i,\btheta}-f_{i,\btheta'}\|_\infty
        \leqslant\gamma_i\|\btheta-\btheta'\|_2,
        \qquad \forall\btheta,\btheta'\in\Theta_i,
    \end{align}
    for some \(\gamma_i>0\). Define \(\log_+
    (u):=\max\{0,\log u\}\) for \(u>0\). With 
    probability at least \(1-\delta\), simultaneously
    for all \(n\geqslant1\),
    \begin{align}\label{eq:fgts_regret_c}
        \mathcal R_{i,n} &\leqslant \frac{m_i\bar\kappa_n 
        }{\eta_i}\lambda_{i,n} n + \frac{\log(3/\delta)
        +d_i^\theta\log_+(2\gamma_i n)}{\lambda_{i,n}}
        + \frac{\bar\lambda_{i,n}}{
        \lambda_{i,n} } + \Big(1+\frac{4\eta_i}
        {n\lambda_{i,n}}\Big) \sqrt{n\log 
        \frac{2n}{\delta}}.
    \end{align}
\end{theorem}

For a tuning scheme analogous to the finite-class case, set
\begin{align}
    \eta_i:=\frac1{4},
    \qquad
    \lambda_{i,n}:=
    \sqrt{\frac{\log(6/\delta)
    +d_i^\theta\big(1+\log_+(2\gamma_i)\big)}{4m_i n}}.
\end{align}
If every agent uses this tuning scheme, then, for each
fixed \(i\in[K]\), with probability at least \(1-\delta\),
the following bound holds simultaneously for all
\(n\geqslant1\):
\begin{align}
    \mathcal R_{i,n}
    &\leqslant
    4\log(3n) \sqrt{n m_i \{\log(6/\delta)
    +d_i^\theta\big(1+\log_+(2\gamma_i)\big)\}}.
\end{align}
Thus, for fixed confidence and model parameters, the
regret bound grows as \(\sqrt n\log n\), as in the
finite-class case. Here the parameter dimension
\(d_i^\theta\) and the Lipschitz constant \(\gamma_i\)
determine the complexity term.

Consider a well-specified linear model with a known
additive baseline:
\begin{align}
    f_{i,\btheta}(\mathbf x,\mathbf c)
    =b_i(\mathbf x,\mathbf c)
    +\langle\btheta,\boldsymbol\phi_i(\mathbf x,\mathbf c)
    \rangle,
    \qquad \btheta\in\mathbb B_1(0)\subseteq\mathbb R^d.
\end{align}
The baseline and features are chosen so that all
predictions lie in \([0,1]\). For example, it suffices
to take \(b_i\equiv1/2\) and
\(\|\boldsymbol\phi_i(\mathbf x,\mathbf c)\|_2
\leqslant1/2\). Since the baseline cancels in
\(f_{i,\btheta}-f_{i,\btheta_i'}\), the prediction
errors admit a linear representation of dimension
at most \(d\). We may therefore take
\(d_i^\theta=m_i=d\), and our bound gives
\begin{align}
    \mathcal R_{i,n}=\widetilde O(d\sqrt n),
\end{align}
where the notation suppresses logarithmic factors. 
This recovers the dimensional dependence established
for FGTS by \citet{zhang2022feel}. In comparison, classical
frequentist analyses of linear Thompson sampling give
bounds of order \(\widetilde O(d^{3/2}\sqrt n)\);
see \citet{abeille2017linear}. The resulting dimension 
dependence improves by a factor \(\sqrt d\) over these 
classical Thompson sampling bounds. This comparison 
concerns frequentist
guarantees for a fixed true parameter and should not be
interpreted as a comparison with Bayesian regret, which
averages over a prior on the true parameter. Our analysis
retains the FGTS dimensional dependence while providing
a high-probability guarantee that holds simultaneously
over all visit counts, as required by Meta-Frieren.

For general linear bandits, the rate
\(\widetilde O(d\sqrt n)\) matches, up to logarithmic
factors, the classical worst-case expected-regret lower
bound of order \(d\sqrt n\) in the nontrivial horizon
regime; see \citet{dani2008stochastic}.

\subsection{Regret under Meta-Frieren}\label{sec:fgts_meta}

We now combine the preceding anytime regret
guarantees for FGTS with the general Meta-Frieren
analysis. In both the finite and continuous
well-specified settings, the agents use the
horizon-independent tuning schemes introduced
above and retain all their past observations.

We additionally assume that, at each visit to
index \(i\), the new context has conditional
distribution \(P_i\), given the history preceding
its observation. Under
\Cref{ass:FGTS-N} and \Cref{ass:FGTS-F}, the reward noise is
conditionally \(1\)-sub-Gaussian and
\(f_i^\star(\mathbf C_{i,n})\in[0,1]\).
Hence, the noise and context concentration
requirements of \Cref{assump:regret_terms} hold
with
\begin{align}\label{eq:fgts_meta_rad}
    \mathrm{rad}_i^{\mathrm{noise}}(n,\alpha)
    &:=\sqrt{\frac{2\log(\pi_n/\alpha)}{n}},
    \quad
    \mathrm{rad}_i^{\mathrm{ctx}}(n,\alpha)
    :=\sqrt{\frac{\log(\pi_n/\alpha)}{2n}}.
\end{align}
Fix \(\delta\in(0,1)\) and set
\(\alpha:=\delta/(3K)\). Define
\begin{align}\label{eq:fgts_meta_w}
    w(n):=3\sqrt{2\log(\pi_n/\alpha)}.
\end{align}
Then, for every \(1\leqslant n\leqslant T\), \(
    2\mathrm{rad}_i(n)
    \leqslant \frac{w(T)}{\sqrt n}\). 
We first consider finite parameter spaces. 
Building on the anytime bound~\eqref{eq:regr4}, we  define
\begin{align}
    B_i(n,\alpha)
    :=\frac52\log(6n)
    \sqrt{nm_i\log(2|\Theta_i|/\alpha)}
\end{align}
and set \(u(n) = \max_{i\in[K]} B_i(n,\alpha)/\sqrt{n}\).

\begin{theorem}\label{th:frieren_fgts_finite}
    Suppose that
    \Cref{ass:FGTS-N}, \Cref{ass:FGTS-F} and \Cref{ass:FGTS-P} 
    hold for every index, and that the
    finite-class FGTS agents use the tuning
    scheme~\eqref{eq:choice}, with confidence
    parameter \(\alpha=\delta/(3K)\).
    Assume that each \(\Theta_i\) is finite,
    its prior is uniform, and the model is
    well specified. 
    Then, with probability at least
    \(1-\delta\), simultaneously for all
    \(T\geqslant K\), Meta-Frieren satisfies
    \begin{align}\label{eq:fgts_finite_meta_regret}
        \mathcal R(T)
        &\leqslant
        u(T)\sqrt{T|\mathcal I^\star|}
        +\big(2u(T) + w(T)\big)^2
        \sum_{i\notin\mathcal I^\star}
        \frac1{\Delta_i}
        +(K-|\mathcal I^\star|)u(T)
        +\|\boldsymbol\Delta\|_1.
    \end{align}
\end{theorem}

We next consider the continuous parameter
spaces of \Cref{th:fgts_cont}. Here the
complexity of each class depends on its
parameter dimension and the Lipschitz
constant of its predictions. Define
\begin{align}\label{eq:fgts_parametric_meta_B}
    B_i^\theta(n,\alpha)
    &:=4\log(3n) \sqrt{n m_i \{\log(6/\alpha)
    +d_i^\theta\big(1+\log_+(2\gamma_i)\big)\}}
\end{align}
and set \(u^\theta(n) = \max_{i\in[K]} B_i^\theta(n,\alpha)/\sqrt{n}\).

\begin{theorem}\label{th:frieren_fgts_parametric}
    Suppose that
    \Cref{ass:FGTS-N}, \Cref{ass:FGTS-F} and \Cref{ass:FGTS-P} 
    hold for every index. Assume also
    that, for each \(i\in[K]\), the
    parameter space, prior, realizability,
    and Lipschitz conditions of
    \Cref{th:fgts_cont} are satisfied.
    Each agent uses the horizon-independent
    tuning scheme specified after
    \Cref{th:fgts_cont}, with confidence
    parameter \(\alpha=\delta/(3K)\). 
    Then, with probability at least
    \(1-\delta\), simultaneously for all
    \(T\geqslant K\), Meta-Frieren satisfies
    \begin{align}\label{eq:fgts_parametric_meta_regret}
        \mathcal R(T)
        &\leqslant
        u^\theta(T)\sqrt{T|\mathcal I^\star|}
        +\big(2
        u^\theta(T) + w(T)\big)^2
        \sum_{i\notin\mathcal I^\star}
        \frac1{\Delta_i}
        +(K-|\mathcal I^\star|)
        u^\theta(T)
        +\|\boldsymbol\Delta\|_1.
    \end{align}
\end{theorem}

The proofs are given in
Appendix~\ref{sec:frieren_fgts}.
Both bounds have the same structure:
the first term controls learning within
optimal indices, while the gap-dependent
term accounts for the cost of selecting
and learning within suboptimal indices.
The remaining terms reflect the initial
exploration of the candidate indices.

For fixed confidence level and model
parameters, the quantities \(u(T)\),
\(u^\theta(T)\), and
\(w(T)\) grow at most logarithmically
in \(T\). Consequently, both theorems give
\begin{align}
    \mathcal R(T)
    =\widetilde O\Big(
        \sqrt{T|\mathcal I^\star|}
        +\sum_{i\notin\mathcal I^\star}
        \frac1{\Delta_i}
        +K+\|\boldsymbol\Delta\|_1
    \Big).
\end{align}
Thus, as for GP-UCB agents, the
meta-level selection task preserves
the square-root dependence on the
horizon, up to logarithmic factors.

\color{black}
\section{Experiments}\label{sec:experiments}

\begin{figure}[t]
    \centering
    \vspace*{-15pt}
    \subfloat[\tiny FGTS Experiment]{%
        \label{fig:fgts_exp}%
        \centering
        \includegraphics[width=0.3\textwidth]{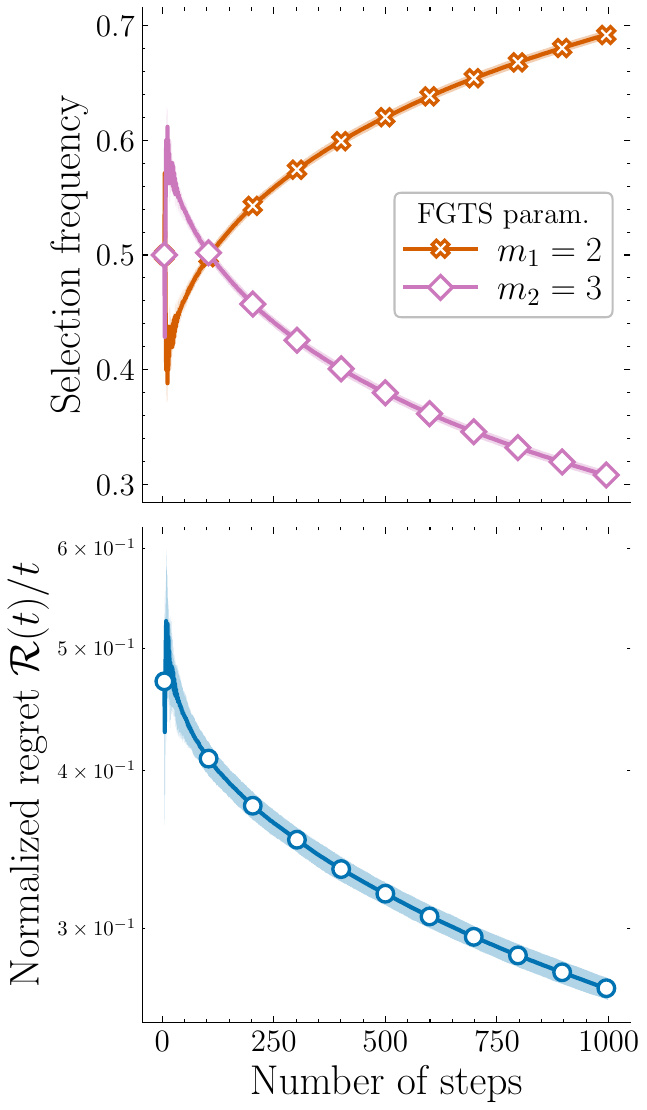}
    }
    \subfloat[\tiny Frequentist GP-UCB]{%
        \label{fig:freq_gp_corr}%
        \centering
        \includegraphics[width= 0.3\textwidth ]{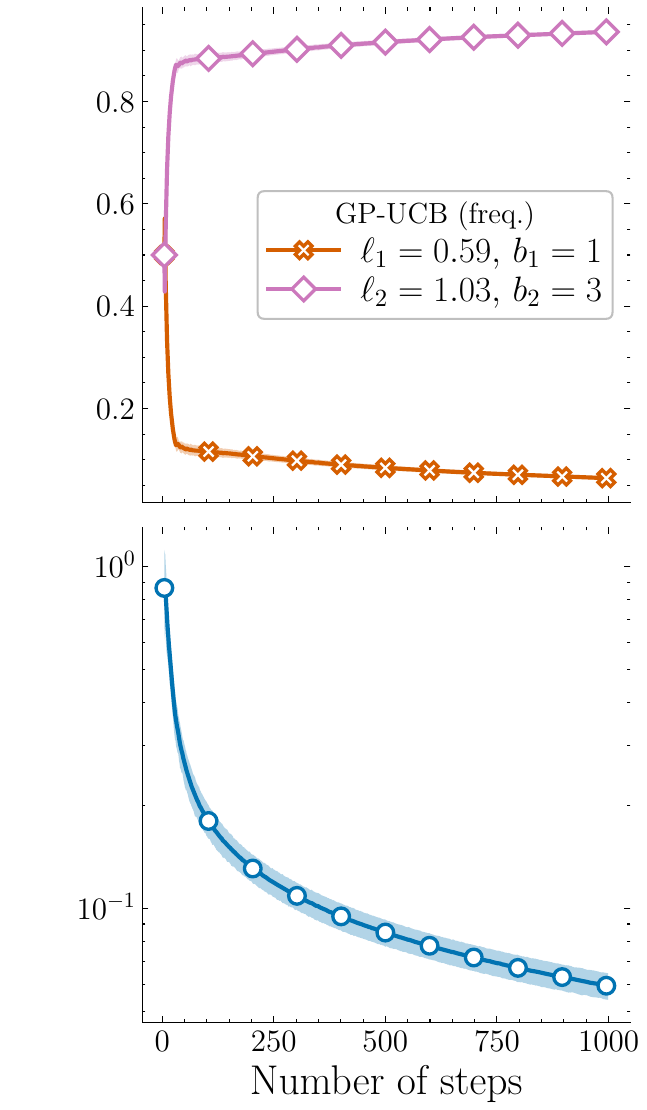}
    }
    \subfloat[\tiny Frequentist GP-UCB,\\ \ \ \ \ misspecified]{%
        \label{fig:freq_gp_missp}%
        \centering
        \includegraphics[width= 0.3\textwidth ]{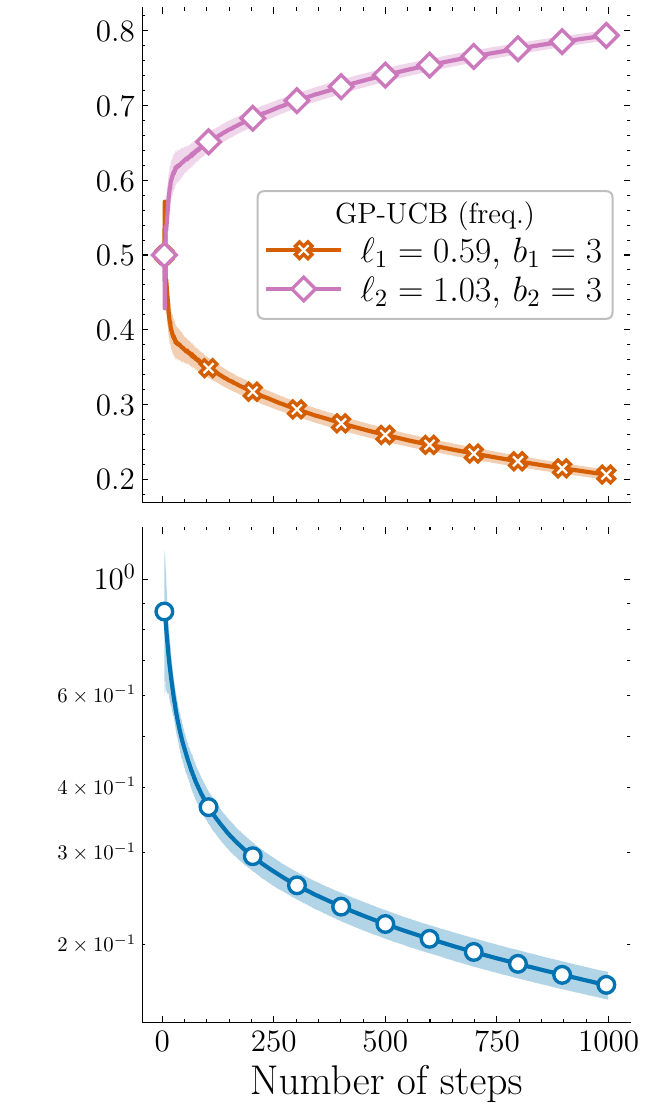}
    }
  
    \caption{Top: frequency of selecting the 
    corresponding index. Bottom: 
    time-normalized regret \(\mathcal{R}(t)/t\).} 
    \vspace*{-5pt}
    \label{fig:big}
\end{figure}

We evaluate Meta-Frieren under the GP-UCB and FGTS settings analysed in \Cref{sec:gp-ucb} and \Cref{sec:fgts}. 
In each experiment, we report the time-normalized  
regret \({\mathcal{R}(t)}/{t}\) at round \(t\), as well as the 
fraction of times each index has been selected up to round \(t\). 
All experiments are conducted for \(T=1000\) rounds using 64 distinct 
seed values. We report averages together with 1.96 standard-error bars.

Each experiment considers a two-index setting where \(\mathbf{C}_t\) is 
sampled independently and uniformly from the interval \([-1, 1]\) for every \(t\in[T]\). 
The action variables are one-dimensional and take values in this interval for all indices.

\paragraph{FGTS experiment} We consider the following linearly embeddable reward functions:
\begin{align}
f_i(x, c) = \varphi_i(x,c)^\top\phi_i(x)/m_i,
\end{align}
where \(\varphi_i: \mathbb{R}^2\to \mathbb{R}^{m_i}\) and \(\phi_i: \mathbb{R}\to \mathbb{R}^{m_i}\) 
are affine maps composed with a coordinate-wise sigmoid function, for all \(i\in\{1,2\}\). The parameter space \(\Theta_i\) is defined
by varying the affine parameters of \(\varphi_i\) over \(\mathbb{R}\). In our experiment, we follow the specification of \Cref{th:fgts_cont} 
with \(m_1=2\) and \(m_2=3\). The parameters of the reward functions are chosen such that \(\mu_1=0.8\) and \(\mu_2=0.1\).

\begin{figure}[t]
    \centering
    \vspace*{-15pt}
    \subfloat[\tiny Bayesian GP-UCB]{%
        \label{fig:by_gp_ucb}%
        \centering
        \includegraphics[width=0.3\textwidth]{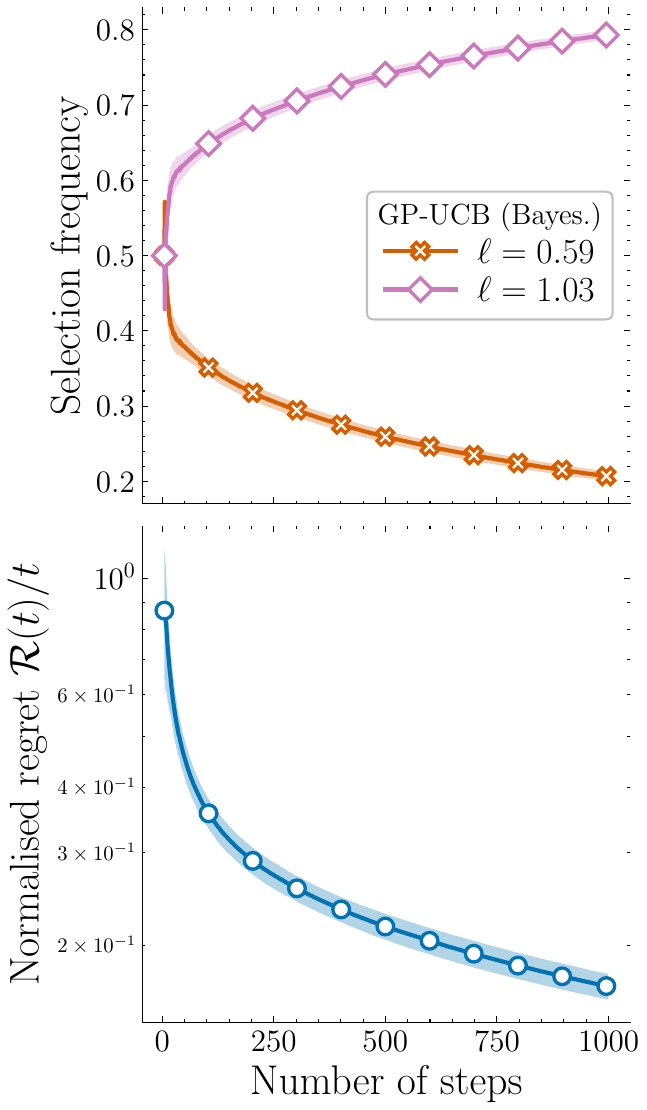}
    }
    \subfloat[\tiny SCM: CoCa-BO]{%
        \label{fig:cocabo_them}%
        \centering
        \includegraphics[width= 0.3\textwidth ]{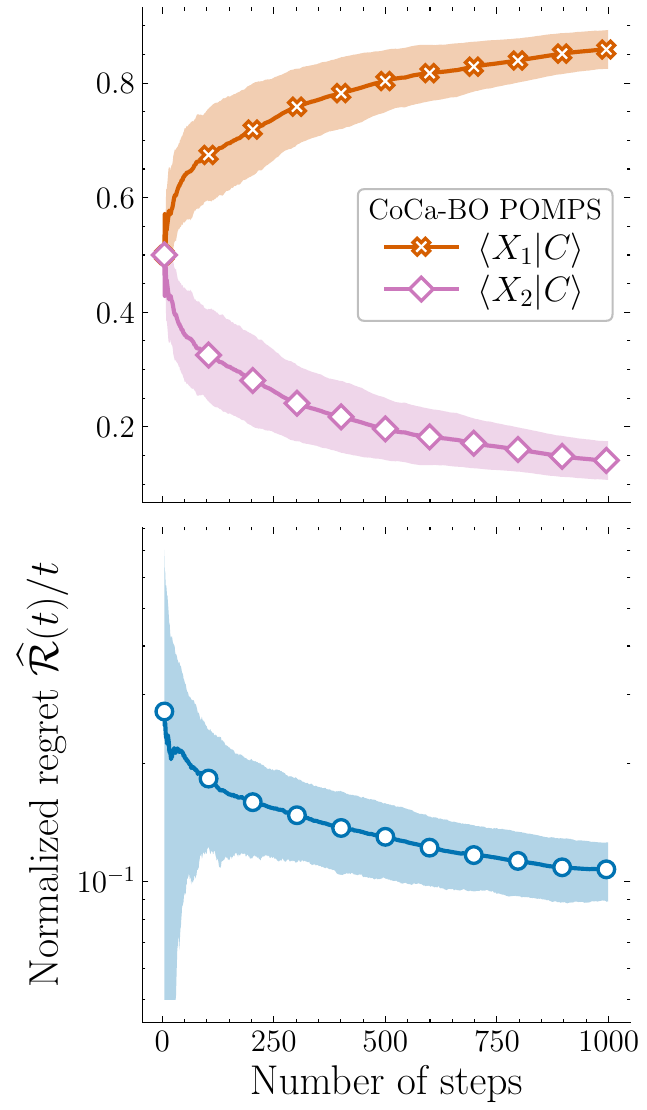}
    }
    \subfloat[\tiny SCM: Meta-Frieren]{%
        \label{fig:cocabo_us}%
        \centering
        \includegraphics[width= 0.3\textwidth ]{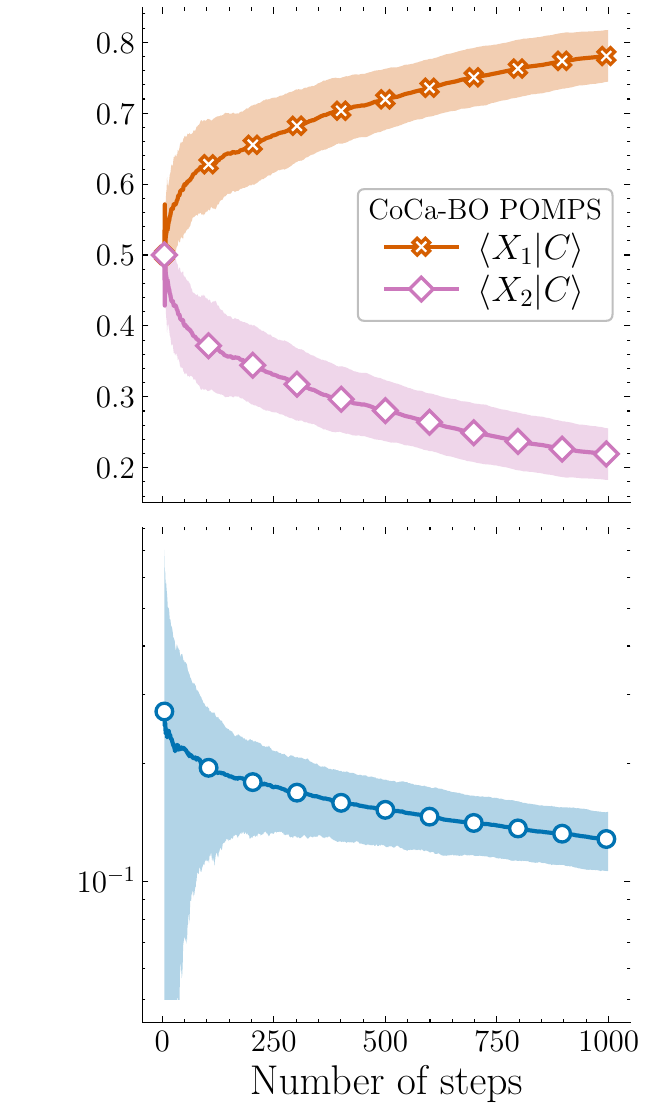}
    }

    \caption{The first column reports \(\mathcal{R}(t)/t\). The second and third columns
    report \(\widehat{\mathcal R}(t)/t\), where \(\widehat{\mathcal R}(t)\) is the regret definition of CoCa-BO~\eqref{eq:regret}.} 
    \vspace*{-5pt}
    \label{fig:big2}
\end{figure}

As shown in \Cref{fig:fgts_exp}, Meta-Frieren initially selects the suboptimal index more frequently. This behaviour 
is expected because \(m_2>m_1\) gives the second index a larger exploration bonus in~\eqref{eq:fgts_parametric_meta_B}. After 
approximately 125 rounds, Meta-Frieren begins to favour the optimal index, and its selection frequency increases thereafter. 
At the same time, the time-normalized regret decreases, which is consistent with sublinear regret. 
The implementation details are provided in Appendix~\ref{app:fgts_exp_det}.

\paragraph{GP-UCB experiments} In this set of experiments, the \(i^{th}\) index is assigned the reward function
\begin{align}\label{eq:gp_ucb_exp}
f_i(x, c) = (x - c)^2\exp\Big(-\frac{x^2 + c^2}{2\ell_i^2}\Big),
\end{align}
with lengthscale \(\ell_i > 0\). We set \(\ell_1=2^{-3/4}\) and \(\ell_2=\sqrt{3/(2\sqrt{2})}\),
which yield \(\mu_1\approx0.443\) and \(\mu_2\approx1.194\).
We defer the implementation details for both configurations to
Appendix~\ref{app:gp_ucb_exp_det}.

\subparagraph{Frequentist configuration} 
We use \(b_1\) and \(b_2\) in~\eqref{eq:b_gpucb_rad1},~\eqref{eq:b_gpucb_B1} instead of a general 
bound \(b\), without invalidating our theory in \Cref{sec:freq_gp_ucb}.
Our choices of \(\ell_1\) and \(\ell_2\) ensure that 
\(f_1\) and \(f_2\) have bounded RKHS norms under squared-exponential kernels with the 
corresponding lengthscales. More precisely, 
\(\lVert f_1\rVert_{k_1}=1 \) and 
\(\lVert f_2\rVert_{k_2}=3\), where 
\(k_i\) is a 
squared-exponential kernel with lengthscale \(\ell_i\) on the joint action-context domain
(see Appendix~\ref{app:rkhs_norm} for the calculation).

\Cref{fig:freq_gp_corr} shows that Meta-Frieren rapidly concentrates its selections
on the optimal index and exhibits sublinear regret. We additionally consider 
a setting with \(b_1=b_2=3\), thereby eliminating the additional exploration 
bonus associated with a larger RKHS norm bound. We use a Matérn-\(3/2\) 
kernel whose hyperparameters are selected by maximizing the log marginal likelihood. 
As shown in \Cref{fig:freq_gp_missp}, Meta-Frieren continues to identify the optimal 
index despite the misspecified kernel and estimated lengthscales.

\subparagraph{Bayesian configuration}
We consider the same reward functions as in~\eqref{eq:gp_ucb_exp}.
The action and context dimensions and domain widths are set to
\(d_i^x=d_i^c=1\) and \(w_i^x=w_i^c=2\) in~\eqref{eq:b_gpucb_rad}, respectively, for every
\(i\). We again use a Matérn-\(3/2\) kernel and estimate its lengthscale by
maximizing the log marginal likelihood. The results are presented in
\Cref{fig:by_gp_ucb}.

\paragraph{Comparison with CoCa-BO}
We compare Meta-Frieren with CoCa-BO using the Bayesian GP-UCB implementation of \citet{arsenyan2026}.
The two methods differ only in their index selection rules: Meta-Frieren uses meta-UCB~\eqref{eq:ucb_def},
whereas CoCa-BO uses the running average of the UCB scores evaluated at the previously selected points for each index.
We consider the structural causal model (SCM) from the second configuration in Section~6 of~\citet{arsenyan2026},
which consists of two indices sharing the same contextual variable \(C\), but having distinct action variables,
denoted by \(X_1\) and \(X_2\). In this configuration, the optimal index corresponds to \(X_1\). Further details are provided in \citet{arsenyan2026}.

The results for CoCa-BO and Meta-Frieren are shown in \Cref{fig:cocabo_them,fig:cocabo_us}, respectively. 
In both figures, \(\langle X_1\mid C\rangle\) denotes the first index, whose agent selects \(X_1\) based on 
the observed context \(C\), while \(\langle X_2\mid C\rangle\) denotes the second index. To ensure a 
direct comparison with \citet{arsenyan2026}, we report \(\widehat{\mathcal{R}}(t)/t\) rather than \(\mathcal{R}(t)/t\). 
Comparing the two figures, Meta-Frieren identifies the optimal index and exhibits a regret profile similar to that of CoCa-BO, 
although its concentration on the optimal index is slightly slower.

\begin{table}[t]
    \centering
    \caption{Hyperparameter search spaces for the three model classes tuned on the MAGIC dataset.
    Closed intervals denote continuous ranges, while braces list discrete values.}
    \label{tab:magic_hpo_spaces}
    \begin{tabular}{lll}
    \toprule
    Model class & Hyperparameter & Range \\
    \midrule
    GBDT & learning rate & \([10^{-3},\, 1]\) \\
         & max.\ number of leaf nodes & \(\{8, \dots, 128\}\) \\
         & min.\ number of samples per leaf & \(\{5, \dots, 200\}\) \\
         & \(\ell_2\) regularization & \([10^{-6},\, 10^{2}]\) \\
    \midrule
    SVM  & penalty parameter \(C\) & \([10^{-3},\, 10^{3}]\) \\
         & kernel coefficient \(\gamma\) & \([10^{-4},\, 10]\) \\
    \midrule
    \(k\)-NN & number of neighbours \(k\) & \(\{1, \dots, 256\}\) \\
         & neighbour weighting & \{uniform, distance\} \\
         & distance metric & \{Manhattan, Euclidean, Chebyshev\} \\
    \bottomrule
    \end{tabular}
    \end{table}

\paragraph{Hyperparameter optimization experiment}

We conclude our experiments with a hierarchical hyperparameter optimization task in which, 
at each round, the algorithm selects a model class and optimizes its hyperparameters. 
We consider the task of discriminating gamma-ray signals from hadronic background using the 
MAGIC Gamma Telescope dataset~\citep{magic_gamma_telescope_159}. Each instance is represented by 
10 numerical features describing the shape and orientation of the shower image. The dataset 
contains 19,020 instances, of which 12,332 correspond to gamma-ray signals and 6,688 to 
hadronic background. We randomly partition the data into training and validation sets and 
keep this split fixed across all evaluations, following~\citet{eggensperger2021hpobench}.

We consider three model classes: gradient-boosted decision trees (GBDT)~\citep{friedman2001greedy} 
with histogram-binned splitting~\citep{ke2017lightgbm}, a support vector machine (SVM)~\citep{cortes1995support} 
with a Gaussian kernel,\footnote{For the SVM, we follow the parameterization of the \texttt{SVC} class in 
scikit-learn, where \(C\) is the penalty parameter and \(\gamma\) is the coefficient of the Gaussian kernel \(\exp(-\gamma \|\mathbf x-\mathbf x'\|_2^2)\); 
see \url{https://scikit-learn.org/stable/modules/generated/sklearn.svm.SVC.html}.} and \(k\)-nearest neighbours 
(\(k\)-NN)~\citep{dudani1976distance}. The hyperparameter spaces considered for these models are given in 
Table~\ref{tab:magic_hpo_spaces}. We describe the handling of integer and
categorical hyperparameters in Appendix~\ref{app:magic_variable_types}.

\begin{figure}[t]
    \centering
    \vspace*{-15pt}
    \subfloat[\tiny Bayesian GP-UCB]{%
        \label{fig:magic_gp_ucb}%
        \centering
        \includegraphics[width=0.3\textwidth]{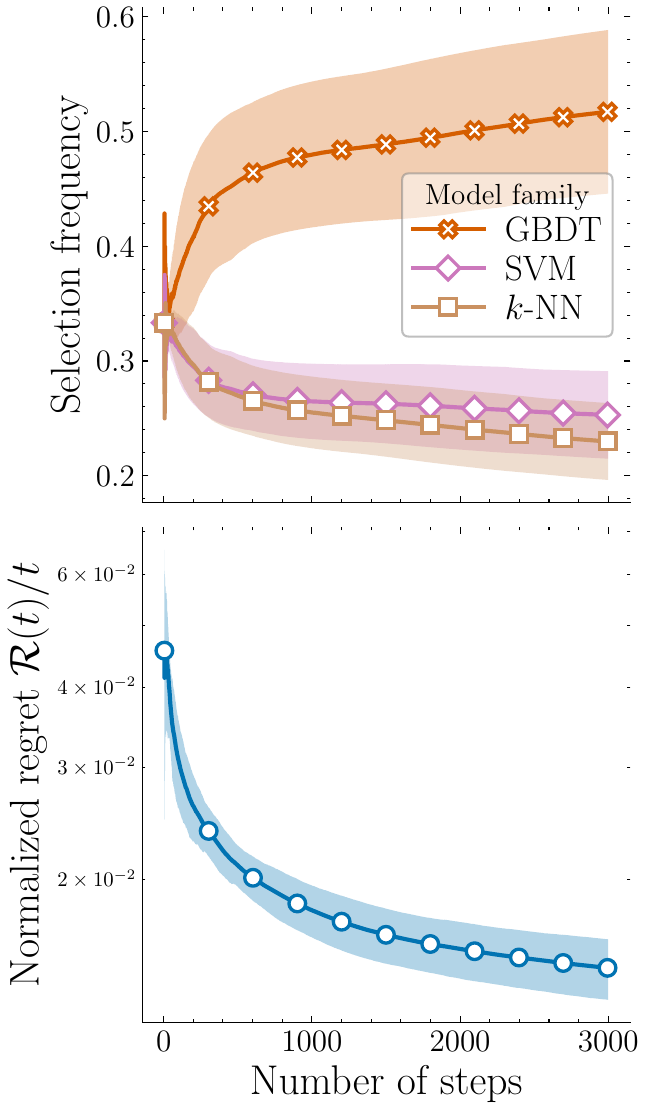}
    }
    \subfloat[\tiny FGTS]{%
    \label{fig:magic_fgts}%
    \centering
    \includegraphics[width= 0.3\textwidth ]{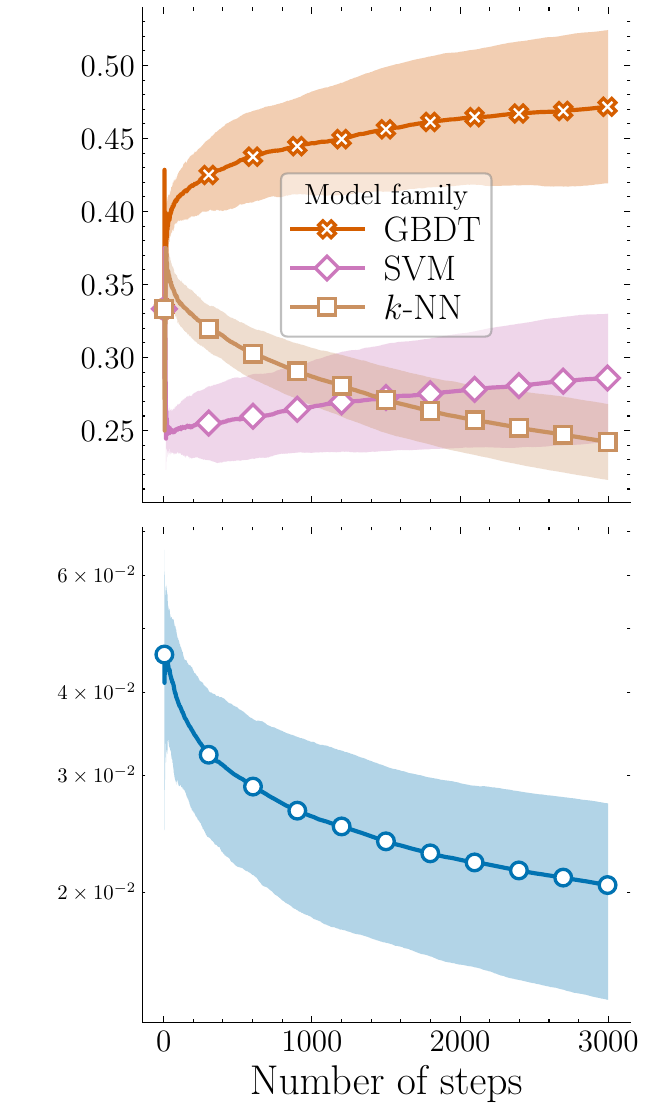}
    }
    \subfloat[\tiny Greedy method]{%
        \label{fig:magic_greedy}%
        \centering
        \includegraphics[width= 0.3\textwidth ]{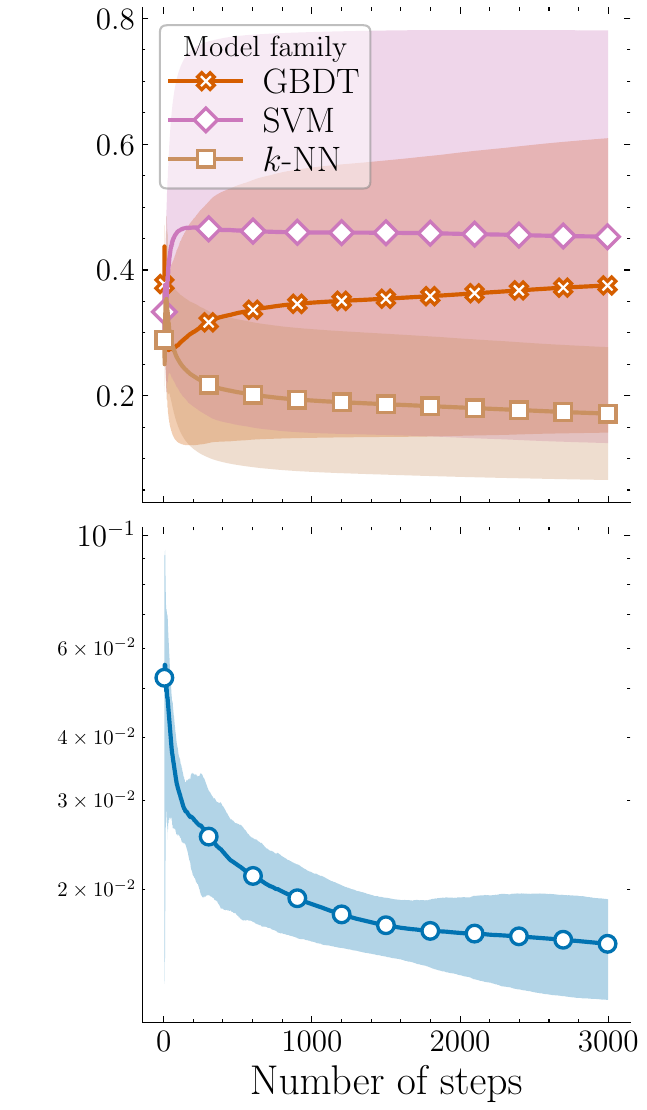}
    }

    \caption{Hierarchical hyperparameter optimization on the MAGIC dataset with the
    Bayesian GP-UCB configuration (a), the FGTS configuration (b), and the greedy
    baseline (c). Each panel shows the selection frequency of each model class (top)
    and the time-normalized regret \(\mathcal{R}(t)/t\) (bottom) over the search
    spaces of Table~\ref{tab:magic_hpo_spaces}.}
    \label{fig:magic_hpo}
\end{figure}

For each model class, we optimize its hyperparameters for 3000 iterations using Bayesian GP-UCB 
and take the highest observed validation area under the curve (AUC) as \(\mu_i\). This procedure 
yields \(\mu_i\) values of \(0.94\), \(0.93\), and \(0.91\) for GBDT, SVM, and \(k\)-NN, respectively.
At each round \(t\), Meta-Frieren selects a model class and a corresponding hyperparameter configuration, 
trains the resulting model, and receives its validation AUC as the reward. We compare the Bayesian 
GP-UCB and FGTS configurations of Meta-Frieren with a greedy baseline. The baseline uses Bayesian 
GP-UCB to select hyperparameters within each model class, but selects the index solely according to 
its empirical mean reward, rather than the meta-UCB criterion in~\eqref{eq:ucb_def}. 
\Cref{fig:magic_hpo} reports the results.

Under the Bayesian GP-UCB configuration (\Cref{fig:magic_gp_ucb}), Meta-Frieren identifies GBDT
as the optimal model class, and its selection frequency increases throughout the run. At the same
time, the time-normalized regret decreases, which indicates that the within-index agent also
locates well-performing hyperparameter configurations for GBDT. Under the FGTS configuration
(\Cref{fig:magic_fgts}), Meta-Frieren likewise selects GBDT increasingly often, although the
selection frequency of SVM also grows. Since the selection frequency of GBDT grows at a faster
rate than that of SVM, we expect GBDT to dominate the selections over a longer horizon. Finally,
the greedy baseline (\Cref{fig:magic_greedy}) performs worse than both configurations of
Meta-Frieren. In \(14\) out of the \(64\) seeds, the empirical mean of the AUC values observed
for SVM during the warm-up rounds exceeds that of GBDT, and the greedy rule subsequently remains
stuck on SVM, so the selection frequency of GBDT cannot exceed \(50/64\). This failure
illustrates the value of the exploration bonus in the meta-UCB criterion~\eqref{eq:ucb_def},
which enables Meta-Frieren to recover from unfavourable initial observations instead of
committing prematurely to a suboptimal model class.

\section*{Conclusion}

We studied a two-stage online optimization problem in which a meta-learner repeatedly selects
an agent-environment pair, after which the chosen agent observes a context, acts in its
environment, and receives a reward. We introduced Meta-Frieren, a general framework for this
class of problems, which requires from each agent only a high-probability bound on its
within-index regret, together with concentration of the optimal contextual rewards and of the
reward noise
(\Cref{assump:regret_terms}).
Under these conditions, \Cref{theor:meta_regret} shows that the meta-UCB
criterion~\eqref{eq:ucb_def} selects each suboptimal index at most \(n_i\left(T\right)\) times
and yields a regret bound driven entirely by the within-index bounds \(B_i\) and the
concentration radii \(\mathrm{rad}_i\).

Since Meta-Frieren does not constrain how agents select their actions, it accommodates
UCB-based methods as well as posterior sampling ones. We instantiated the framework with
GP-UCB agents in both the Bayesian and frequentist settings (\Cref{sec:gp-ucb}),
recovering, in the Bayesian case, the guarantee of CoCa-BO~\citep{arsenyan2026} up to constant
factors, and with FGTS agents (\Cref{sec:fgts}). To enable the latter, we derived, to our
knowledge, the first high-probability regret bound for FGTS (\Cref{prop:fgts_bound}), together
with theorems for realizable finite (\Cref{th:fgts_discrete}) and parametric function classes (\Cref{th:fgts_cont}), the former being
minimax-optimal up to logarithmic factors, and combined them with the meta-level analysis in
\Cref{th:frieren_fgts_finite,th:frieren_fgts_parametric}. Our experiments
support the theory, showing that Meta-Frieren concentrates its selections on the optimal index
and exhibits sublinear regret in both the GP-UCB and FGTS settings, remains effective under a
misspecified kernel, and displays a regret profile comparable to that of CoCa-BO while
applying to a broader class of agents.

\section*{Acknowledgements} 
This work was supported by Hi! PARIS and received government 
funding managed by the Agence Nationale de la Recherche under 
the France 2030 program, references (ANR-23-IACL-0005) and
(ANR-23-PEIA-0004), 
as well as 
by ERC grant SAGMOS (grant agreement No. 101201229).

\bibliography{bibliography}

\newpage

\appendix

\etocdepthtag.toc{appendices}

\begingroup
\etocsettagdepth{main}{none}
\etocsettagdepth{appendices}{subsection}
\etocsettocstyle{\section*{Contents of the appendices}}{}
\tableofcontents
\endgroup

\section{Proofs of Bounds for Meta-Frieren with GP-UCB} 

This section provides proofs of regret upper bounds for Bayesian GP-UCB (\Cref{sec:by_gp_ucb}) and
frequentist GP-UCB (\Cref{sec:freq_gp_ucb}).

\subsection{Proof of \Cref{th:bay_gp_ucb}}\label{sec:gp_ucb_proof}

We begin by verifying that the three conditions
of \Cref{assump:regret_terms} hold with the
bounds specified in~\eqref{eq:b_gpucb_rad}--\eqref{eq:b_gpucb_B}.
Recall that, for each index \(i\), the
observations are numbered according to the
successive visits to that index, rather than
the global rounds of Meta-Frieren.

First, since the observation noises are
independent and distributed as
\(\mathcal N(0,\sigma^2)\), Gaussian
concentration yields
\begin{align*}
    \mathbb P\Big(
    \Big|
    \frac1n\sum_{j=1}^n\varepsilon_{i,j}
    \Big|
    \leqslant
    \mathrm{rad}_i^{\mathrm{noise}}(n,\alpha),
    \quad \forall n\geqslant1
    \Big)
    \geqslant 1-\alpha.
\end{align*}
This establishes \eqref{eq:RwN}.

Second, the optimal contextual rewards
\(f_i^\star(\mathbf C_{i,j})\) are iid
conditionally on \(f_i\). Under the
sample-path regularity condition
\eqref{eq:grad_op_norm_f}, Lemma~3 of
\citep{arsenyan2026} provides the
corresponding uniform concentration
inequality, with radius
\(\mathrm{rad}_i^{\mathrm{ctx}}(n,\alpha)\).
Thus, \eqref{eq:OCR} holds simultaneously
for all \(n\geqslant1\), with probability
at least \(1-\alpha\).

Third, by Theorem~1 of
\citet{krause2011contextual}, the
contextual GP-UCB rule
\eqref{eq:gp_ucb_action}, with exploration
parameter \eqref{eq:beta_t_gp_ucb},
satisfies
\begin{align*}
    \mathbb P\left(
        \mathcal R_{i,n}
        \leqslant B_i(n,\alpha),
        \quad \forall n\geqslant1
    \right)
    \geqslant 1-\alpha.
\end{align*}
This establishes \eqref{eq:ALL}.

Each of these three guarantees concerns
the sequence of observations collected
at a fixed index and holds uniformly
over its local visit counts. They
therefore remain valid when the indices
are selected adaptively by Meta-Frieren. 
Finally, applying a union bound over
the three guarantees and all \(K\)
indices, and recalling that
\(\alpha=\delta/(3K)\), we obtain
\begin{align*}
    \mathbb P\big(\mathcal E(\delta)\big)
    \geqslant 1-3K\alpha
    =1-\delta.
\end{align*}
On this event, the conditions of
\Cref{theor:meta_regret} are satisfied,
and we can apply the general
meta-regret bound.

Let \(n_i:=n_i\left(T\right)\) then, with a slight abuse of notation, we have that 
\begin{equation}
\Delta_i\leqslant \frac{\sqrt{{\mathsf C_\sigma}\beta_i\left(n_i\right)\gamma_{i}\left(n_i\right)}  + \rho_i\left(n_i\right)}{\sqrt{n_i-1}}+\frac{2}{n_i-1},
\end{equation}
where we have used the fact that \(\beta_i\), \(\gamma_i\), and \(\rho_i\) 
are monotonically increasing with \(n_i\). Let \(c\left(T\right)=\sqrt{{\mathsf C_\sigma}\beta\left(T\right)\gamma\left(T\right)}+\rho\left(T\right)\), then using Young's inequality 
\begin{equation}\label{eq:sqrt_play_time}
    (n_i-1)\Delta_i\leqslant {c\left(T\right)}\sqrt{n_i-1}+2\leqslant \frac{(n_i-1)\Delta_i}{2}+\frac{c(T)^2}{2\Delta_i}+2.
\end{equation}
Thus, \((n_i-1)\Delta_i\leqslant \frac{c(T)^2}{\Delta_i}+4\) and if \(n_i=1\), then 
\begin{equation}
    n_i\Delta_i+B_i\left(n_i\right)\leqslant  \Delta_i+\sqrt{{\mathsf C_\sigma}\beta\left(1\right)\gamma\left(1\right)}+2\leqslant  \Delta_i+2\left(1+\sqrt{\beta\left(1\right)}\right),
\end{equation}
as \({\mathsf C_\sigma}\gamma\left(1\right)\leqslant 4\). While for \(n_i\geqslant 2\)
\begin{align}
    B_i\left(n_i\right)&\leqslant \sqrt{{\mathsf C_\sigma}\beta(T)\gamma(T)n_i}+2\leqslant
    \sqrt{2{\mathsf C_\sigma}\beta(T)\gamma(T)(n_i-1)}+2\\
    &\leqslant \frac{(n_i-1)\Delta_i}{2}+\frac{{\mathsf C_\sigma}\beta(T)\gamma(T)}{\Delta_i}+2
\end{align}
Combining the last display with the upper bound on \((n_i-1)\Delta_i\) yields
\begin{align}
    n_i\Delta_i+B_i\left(n_i\right)\leqslant \Delta_i + \frac{4\left({\mathsf C_\sigma}\beta(T)\gamma(T)+\rho^2(T)\right)}{\Delta_i}+8,
\end{align}
when \(n_i\geqslant2\). Noting that \(\beta(1)\geqslant4\), and hence \(2\left(2+\sqrt{\beta\left(1\right)}\right)\geqslant8\) gives
\begin{align}\label{eq:gpucb:meta_proof:bayes:suboptimal}
    n_i\Delta_i+B_i\left(n_i\right)\leqslant \Delta_i + \frac{4\left({\mathsf C_\sigma}\beta(T)\gamma(T)+\rho^2(T)\right)}{\Delta_i}+2\left(2+\sqrt{\beta\left(1\right)}\right),
\end{align}
for all \(n_i\geqslant1\). Finally, for the optimal indices 
\begin{align}
    \max_{\substack{u_i\in\{0\}\cup[T]\\ \sum_{i\in\mathcal{I}^\star} u_i \leqslant T}}
\ \sum_{i\in\mathcal{I}^\star} B_i\left(u_i\right)&\leqslant  2|\mathcal I^\star|+\sqrt{{\mathsf C_\sigma}\beta\left(T\right)}\max_{ \sum_{i\in\mathcal{I}^\star} u_i = T}
\ \sum_{i\in\mathcal{I}^\star} \sqrt{u_i\gamma_i\left(u_i\right)}\\ 
&\leqslant  2|\mathcal I^\star|+\sqrt{|\mathcal{I}^\star|T{\mathsf C_\sigma}\beta\left(T\right)\bar{\gamma}\left(T\right)}. \label{eq:p3}
\end{align}
Combining~\eqref{eq:gpucb:meta_proof:bayes:suboptimal} and~\eqref{eq:p3}, we get that the regret of Meta-Frieren is upper bounded by
\begin{align}
    \mathcal{R}\left(T\right)&\leqslant  \sqrt{|\mathcal{I}^\star|T{\mathsf C_\sigma}\beta\left(T\right)\bar{\gamma}\left(T\right)}+
    4\left({\mathsf C_\sigma}\beta(T)\gamma(T)+\rho^2(T)\right)\sum_{i\notin\mathcal I^\star}\frac{1}{\Delta_i}\\
    &+\|\boldsymbol{\Delta}\|_1+2K\big(2+\sqrt{\beta\left(1\right)}\big).
\end{align}

\subsection{Proof of \Cref{th:freq_gp_ucb}}\label{sec:proof_of_freq_gp_ucb}

We first verify that the radii in~\eqref{eq:b_gpucb_rad1}
satisfy~\eqref{eq:RwN} and~\eqref{eq:OCR}.
The noise radius satisfies~\eqref{eq:RwN} by the
sub-Gaussian noise
condition~\eqref{eq:freq_set:sub_gaussian_noise}.
For the context radius,~\eqref{eq:freq_function_bound}
gives \(f_i^\star(\mathbf c)\in[-b,b]\).
Since the contexts are iid\, Hoeffding's inequality
bounds the probability that the deviation in~\eqref{eq:OCR}
exceeds \(\mathrm{rad}^{\mathrm{ctx}}_i(n,\alpha)\) by
\(2\alpha/\pi_n\) for each fixed \(n\in\mathbb N^+\).
A union bound over \(n\), using
\(\sum_{n\geqslant1}2/\pi_n=1\), then yields~\eqref{eq:OCR}.

\begin{proposition}\label{prop:cgp-regret}
    Under the setting of~\Cref{sec:freq_gp_ucb}, for every \(i\in[K]\),
    \begin{align}
    \label{eq:cgp-regret}
        \mathcal R_{i,n}
        \leqslant
        \sqrt{{\mathsf C_\sigma}\,n\beta_i(n,\alpha)\gamma_{i,n}}
    \end{align}
    with probability at least \(1-\alpha\), simultaneously for every
    \(n\in\mathbb N^+\).
\end{proposition}

\begin{proof}     
Fix \(i\in[K]\) and \(\alpha\in(0,1)\).
Applying Theorem~1 of
\citet{durand2018streaming} with
regularization parameter \(\lambda=\sigma^2\)
and noise sub-Gaussian parameter \(\sigma\),
and using \(\|f_i\|_{k_i}\leqslant b\),
we obtain, with probability at least
\(1-\alpha\), simultaneously for all
\(j\geqslant1\) and
\(\mathbf z\in\mathfrak X_i\times\mathfrak C_i\),
\begin{align}\label{eq:cgp-confidence}
    |f_i(\mathbf z)-m_{i,j-1}(\mathbf z)|
    \leqslant
    \sqrt{\beta_i(j,\alpha)}
    \,s_{i,j-1}(\mathbf z).
\end{align}
Here we have used the fact that, for
\(\lambda=\sigma^2\), the uncertainty
term in the cited theorem coincides
with the GP posterior standard deviation,
together with the monotonicity of
\(\gamma_{i,j}\) in \(j\). 
    On this event,
    the GP-UCB action selection rule and \eqref{eq:cgp-confidence} give
    \begin{align}
        f_i^\star(\mathbf C_{i,j})-f_i(\mathbf Z_{i,j})
        &\leqslant
        m_{i,j-1}(\mathbf Z_{i,j})
        +\sqrt{\beta_i(j,\alpha)}\,s_j-f_i(\mathbf Z_{i,j})\leqslant 2\sqrt{\beta_i(j,\alpha)}\,s_j,
    \end{align}
    where \(\mathbf Z_{i,j}:=(\mathbf X_{i,j},\mathbf C_{i,j})\) and
    \(s_j:=s_{i,j-1}(\mathbf Z_{i,j})\).
    Let \(\mathbf K_{i,n}
    :=(k_i(\mathbf Z_{i,j},\mathbf Z_{i,\ell}))_{j,\ell=1}^n\).
    \citet[Lemma~5.3]{srinivas2009gaussian} yields
    \begin{align*}
        \sum_{j=1}^n\log(1+\sigma^{-2}s_j^2)
        =\log\det(\mathbf I+\sigma^{-2}\mathbf K_{i,n})
        \leqslant 2\gamma_{i,n}.
    \end{align*}
    Since \(0\leqslant s_j^2\leqslant
    k_i(\mathbf Z_{i,j},\mathbf Z_{i,j})\leqslant1\), concavity
    of \(u\mapsto\log(1+\sigma^{-2}u)\) implies
    \begin{align*}
        \sum_{j=1}^n s_j^2
        \leqslant
        \frac{\sum_{j=1}^n\log(1+\sigma^{-2}s_j^2)}
             {\log(1+\sigma^{-2})}
        \leqslant
        \frac{2\gamma_{i,n}}{\log(1+\sigma^{-2})}.
    \end{align*}
    Consequently, monotonicity of \(\beta_i(\cdot,\alpha)\)
    and Cauchy-Schwarz give
    \begin{align}
        \mathcal R_{i,n}
        \leqslant
        2\sqrt{\beta_i(n,\alpha)}\sum_{j=1}^n s_j\leqslant\sqrt{{\mathsf C_\sigma}n\beta_i(n,\alpha)\gamma_{i,n}}.
    \end{align}
This completes the proof of the proposition. 
\end{proof}

We are now ready to conclude the proof of~\Cref{th:freq_gp_ucb}. 
We work on the high-probability event \(\mathcal E(\delta)\).
From \Cref{theor:meta_regret},
\begin{align}
    \sqrt{n_i-1}\leqslant \frac{\sqrt{{\mathsf C_\sigma}\beta(T)\gamma(T)}+\rho(T)}
    {\Delta_i}
\end{align}
Consequently,
\begin{align}
    \Delta_i n_i
    &\leqslant \Delta_i+
    \frac{\big(\sqrt{{\mathsf C_\sigma}\beta(T)\gamma(T)}+\rho(T)\big)^2}
            {\Delta_i}.
\end{align}
Applying the bound on the number of suboptimal-index visits to the within-index regret gives
\begin{align}
    \mathcal R_{i,n_i}\leqslant \mathcal R_{i,n_i-1}+2b\leqslant B_i(n_i-1)+2b
    \leqslant 2b+
    \frac{{\mathsf C_\sigma}\beta(T)\gamma(T)
    +\rho(T)\sqrt{{\mathsf C_\sigma}\beta(T)\gamma(T)}}{\Delta_i}
\end{align}
where the first inequality separates the last visit, whose
regret is at most \(2b\). 
For optimal indices, Cauchy-Schwarz and the definition of
\(\bar\gamma(T)\) yield
\begin{align}
    \sum_{i\in\mathcal I^\star}\mathcal R_{i,N_i(T)}
    \leqslant \sqrt{{\mathsf C_\sigma}\beta(T)}
    \sum_{i\in\mathcal I^\star}
    \sqrt{N_i(T)\gamma_{i,N_i(T)}}\leqslant
    \sqrt{T|\mathcal I^\star|{\mathsf C_\sigma}\beta(T)\bar\gamma(T)}.
\end{align}
Combining these estimates with~\eqref{eq:our_regret} and
applying Young's inequality gives
\begin{align*}
    \mathcal R(T)
    &\leqslant
    \sqrt{T|\mathcal I^\star|{\mathsf C_\sigma}\beta(T)\bar\gamma(T)}+
    4\big({\mathsf C_\sigma}\beta(T)\gamma(T)+\rho^2(T)\big)
    \sum_{i\notin\mathcal I^\star}\frac1{\Delta_i}\\
    &+
    2b\big(K-|\mathcal I^\star|\big)
    +\|\boldsymbol\Delta\|_1,
\end{align*}
on the event \(\mathcal{E}(\delta)\).

\newpage


\section{Regret bounds for Feel-Good Thompson Sampling}\label{sec:fgts_proofs}

This appendix proves the results in \Cref{sec:fgts}.
We first establish a general regret bound that allows
arbitrary reward bounds and noise variance proxies.
We then derive the finite- and continuous-parameter
results stated in the main text.

\subsection{General assumptions and the decoupling 
coefficient}

We first extend the assumptions of the main text to
arbitrary reward bounds and noise variance proxies.
This formulation makes the dependence of the regret
bounds on these scales explicit. The main-body results
follow by setting \(\mathsf A_i=\sigma_i=1\).
When the scales are known, an equivalent normalization
can be obtained by rescaling the rewards and candidate
functions, together with the tuning parameters.

\begingroup
\renewcommand{\theassumption}{(FGTS-N')}
\begin{assumption}\label{ass:FGTS-N'}
    For every \(i\in[K]\), there exists \(\sigma_i
    \geqslant 0\) such that for all \(j\geqslant1\),
    \begin{align}\label{eq:fgts_noise_gen}
        R_{i,j}=f_i(\mathbf X_{i,j},\mathbf C_{i,j}) 
        +\varepsilon_{i,j}, \qquad
        \mathbb E\big[e^{\rho\varepsilon_{i,j}}\bigm|
        H_{\tau_{i,j}-1},\mathbf C_{i,j},\mathbf X_{i,j}
        \big] \leqslant e^{\rho^2\sigma_i^2/2},
        \quad \forall\rho\in\mathbb R.
    \end{align}    
\end{assumption}
\addtocounter{assumption}{-1}
\endgroup
\begingroup
\renewcommand{\theassumption}{(FGTS-F')}
\begin{assumption}\label{ass:FGTS-F'}
    The true and candidate reward functions are 
    deterministic and satisfy the following conditions 
    for every \(i\in[K]\). 
    \begin{enumerate}
        \item  There is a fixed 
        \(\mathsf A_i>0\) such that, for every 
        \(\btheta\in\Theta_i\) and \((\mathbf x,\mathbf
        c)\in\mathfrak X_i\times \mathfrak C_i\),
        \begin{align}\label{eq:fgts_bounded_rewards_gen}
            0\leqslant f_i(\mathbf x,\mathbf c)\leqslant
            \mathsf A_i, \qquad 0\leqslant f_{i,\btheta}
            (\mathbf x,\mathbf c) \leqslant\mathsf A_i.
        \end{align}
        \item There are an integer \(m_i\geqslant1\), 
        measurable functions \(u_{i,1},\ldots,u_{i,m_i}:
        \Theta_i\times\mathfrak C_i\to\mathbb R\), and 
        measurable functions \(\phi_{i,1},\ldots,
        \phi_{i,m_i}: \mathfrak X_i\times\mathfrak C_i\to 
        \mathbb R\) such that
        \begin{align}
            f_{i,\btheta}(\mathbf x,\mathbf c)-f_i(
            \mathbf x,\mathbf c) = \sum_{\ell=1}^{m_i}
            u_{i,\ell}(\btheta,\mathbf c)
            \phi_{i,\ell}(\mathbf x,\mathbf c), \qquad
            (\btheta, \mathbf x,\mathbf c)\in \Theta_i
            \times \mathfrak X_i\times \mathfrak C_i.
        \end{align}
        For each fixed \(\mathbf c\), the functions 
        \(u_{i,\ell}(\cdot,\mathbf c)\) are bounded.
    \end{enumerate}   
\end{assumption}
\addtocounter{assumption}{-1}
\endgroup

The final assumption specifies the tuning conditions
for the FGTS algorithm.

\begingroup
\renewcommand{\theassumption}{(FGTS-P')}
\begin{assumption}\label{ass:FGTS-P'}
    For every \(i\in [K]\),
    \vspace{-2pt}
    \begin{enumerate}\itemsep=0pt
        \item \(\eta_i\) is deterministic, positive, and
        satisfies \(4\eta_i(\sigma_i^2\vee\mathsf A_i^2)
        \leqslant 1\).
        \item The sequence \( ( \lambda_{i,j} )_{j\geqslant
        1}\) is deterministic, positive, and nonincreasing. 
    \end{enumerate}
\end{assumption}
\addtocounter{assumption}{-1}
\endgroup
We follow \cite{zhang2022feel} and introduce the decoupling
coefficient that compares the prediction error of a
candidate at its own greedy action with the prediction error
of an independent candidate evaluated at that action. Define
\begin{align}\label{eq:fgts_error_def}
    e_i(\btheta,\tilde\btheta;\mathbf c)
    :=f_{i,\tilde\btheta}\big(\mathbf x_i(\btheta,\mathbf c),
    \mathbf c\big)
    -f_i\big(\mathbf x_i(\btheta,\mathbf c),\mathbf c\big).
\end{align}
For a measurable subset \(\Theta'\subseteq\Theta_i\), let
\begin{align}\label{eq:fgts_dc_def}
    \dc_i(\Theta')
    :=1\vee \sup_{q:\,q(\Theta')=1} \sup_{\mathbf c\in\mathfrak C_i}
    \frac{\big|\mathbb E_{\bvartheta\sim q}
    [e_i(\bvartheta,\bvartheta;\mathbf c)]\big|}
    {\big(\mathbb E_{(\bvartheta,\tilde\bvartheta)\sim q\otimes q}
    [e_i(\bvartheta,\tilde\bvartheta;\mathbf c)^2]\big)^{1/2}}.
\end{align}
We use the conventions \(0/0:=0\) and \(a/0:=+\infty\)
for \(a>0\). 
This quantity appears naturally in the analysis of the 
regret of FGTS. The goal of the second part of 
\Cref{ass:FGTS-F'}, as shown in the next lemma, is to provide 
a simple upper bound on the decoupling coefficient.

\begin{lemma}
    If the second part of \Cref{ass:FGTS-F'} holds, 
    then \(\dc_i^2(\Theta_i)\leqslant m_i\) for every
    \(i\in[K]\).
\end{lemma}
\begin{proof}
    To ease notation and without any loss of generality, we
    remove the dependence of all the quantities on \(i\) and
    on \(c\). Thus, for instance, we write \(e_{\btheta}(
    \mathbf x)=f_{\btheta}(\mathbf x)-f(\mathbf x)\) instead 
    of \(e_{i,\btheta}(\mathbf x,\mathbf c) = f_{i,\btheta}
    (\mathbf x,\mathbf c)-f_i(\mathbf x, \mathbf c)\).
    Introduce the vector-valued functions
    \begin{align}
        \mathbf u(\btheta)
        &:=
        \begin{bmatrix}
            u_1(\btheta)\\
            \vdots\\
            u_m(\btheta)
        \end{bmatrix},
        \qquad
        \boldsymbol\phi(\mathbf x)
        :=
        \begin{bmatrix}
            \phi_1(\mathbf x)\\
            \vdots\\
            \phi_m(\mathbf x)
        \end{bmatrix}.
    \end{align}
    Then \(e_{\btheta}(\mathbf x) =\langle\mathbf u(\btheta), 
    \boldsymbol\phi(\mathbf x)\rangle\). For a probability 
    measure \(q\) on \(\Theta\), consider the \( m\times 
    m \) matrix \(\mathbf S :=\mathbb E_{\bvartheta\sim q}
    \big[\mathbf u(\bvartheta)\mathbf u(\bvartheta)^\top
    \big]\), which is finite because the functions \(u_\ell\) 
    are bounded. Let \(\mathbf S^\dagger\) denote its 
    Moore-Penrose inverse. If \(\mathbf v\) is a vector from 
    a basis of  \(\mathrm{ker}(\mathbf S)\), then
    \(0 =\mathbf v^\top\mathbf S\mathbf v
        =\mathbb E_{\bvartheta\sim q}\big[
            \langle\mathbf v,\mathbf u(\bvartheta)\rangle^2
        \big]\), and hence
    \(\langle\mathbf v,\mathbf u(\bvartheta)\rangle=0\)
    for \(q\)-almost every \(\bvartheta\). Therefore, \(\mathbf u(\bvartheta)\) belongs to the range of \(\mathbf S\) and
    \begin{align}
        \mathbf u(\bvartheta)
        =\mathbf S^{1/2}(\mathbf S^\dagger)^{1/2}
        \mathbf u(\bvartheta)
    \end{align}
    almost surely. Using this identity and the Cauchy--Schwarz
    inequality, we obtain
    \begin{align}
        \left|\mathbb E_{\bvartheta\sim q}\big[
            e_{\bvartheta}(\mathbf x(\bvartheta))
        \big]
        \right|
        &=
        \left|
        \mathbb E_{\bvartheta\sim q}\big[
            \langle\mathbf u(\bvartheta),
            \boldsymbol\phi(\mathbf x(\bvartheta))\rangle
        \big]
        \right|\\
        &=
        \left|
        \mathbb E_{\bvartheta\sim q}\big[
            \big\langle(\mathbf S^\dagger)^{1/2}
            \mathbf u(\bvartheta),
            \mathbf S^{1/2}
            \boldsymbol\phi(\mathbf x(\bvartheta))\big\rangle
        \big]
        \right|\\
        &\leqslant
        \Big(
            \mathbb E_{\bvartheta\sim q}\big[
                \mathbf u(\bvartheta)^\top
                \mathbf S^\dagger\mathbf u(\bvartheta)
            \big]
        \Big)^{1/2}
        \Big(
            \mathbb E_{\bvartheta\sim q}\big[
                \boldsymbol\phi(\mathbf x(\bvartheta))^\top
                \mathbf S
                \boldsymbol\phi(\mathbf x(\bvartheta))
            \big]
        \Big)^{1/2}.
    \end{align}
    The first expectation equals
    \(\operatorname{rank}(\mathbf S)\). Indeed,
    \begin{align}
        \mathbb E_{\bvartheta\sim q}\big[
            \mathbf u(\bvartheta)^\top
            \mathbf S^\dagger\mathbf u(\bvartheta)
        \big]
        =
        \operatorname{tr}\left(
            \mathbf S^\dagger
            \mathbb E_{\bvartheta\sim q}\big[
                \mathbf u(\bvartheta)
                \mathbf u(\bvartheta)^\top
            \big]
        \right)
        =
        \operatorname{tr}(\mathbf S^\dagger\mathbf S)
        =\operatorname{rank}(\mathbf S)
        \leqslant m.
    \end{align}
    Moreover, the second expectation can be written as
    \begin{align}
        &\mathbb E_{\bvartheta\sim q}\big[
            \boldsymbol\phi(\mathbf x(\bvartheta))^\top
            \mathbf S
            \boldsymbol\phi(\mathbf x(\bvartheta))
        \big] =
        \mathbb E_{(\bvartheta,\tilde\bvartheta)\sim q\otimes q}
        \big[
            e_{\tilde\bvartheta}(
            \mathbf x(\bvartheta))^2
        \big].
    \end{align}
    Consequently,
    \begin{align}
        \left|
        \mathbb E_{\bvartheta\sim q}\big[
            e_{\bvartheta}(\mathbf x(\bvartheta))
        \big]
        \right|^2
        \leqslant m \mathbb E_{(\bvartheta,
        \tilde\bvartheta)\sim q\otimes q}
        \big[
            e_{\tilde\bvartheta}(
            \mathbf x(\bvartheta))^2
        \big].
    \end{align}
    Since \(q\) and the context were arbitrary, this proves
    \(\dc(\Theta)^2\leqslant m\) under the linear
    representation assumption. This completes the proof of the lemma.    
\end{proof}

\subsection{A general high-probability bound}\label{sec:fgts_bound}

Another ingredient naturally appearing in the regret analysis is the log-partition function that, roughly speaking, measures how much prior mass is assigned
to candidates that explain the observations. Define the excess
loss relative to the true reward function by
\begin{align}\label{eq:fgts_delta_L}
    \Delta L_{i,j}(\btheta,\mathbf c,\mathbf x,r)
    :=\eta_i\Big[\big(f_{i,\btheta}(\mathbf x,\mathbf c)-r\big)^2
    -\big(f_i(\mathbf x,\mathbf c)-r\big)^2\Big]
    +\lambda_{i,j}\big(f_i^\star(\mathbf c)-f_{i,\btheta}(\mathbf c)\big).
\end{align}
Write \(\Delta L_{i,j}(\btheta)
:=\Delta L_{i,j}(\btheta,\mathbf C_{i,j},\mathbf X_{i,j},R_{i,j})\)
and introduce
\begin{align}\label{eq:fgts_weights}
    W_{i,n}(\btheta):=\exp\Big(-\sum_{j=1}^n
    \Delta L_{i,j}(\btheta)\Big),
    \qquad \widehat Z_{i,n}:=\log\int_{\Theta_i}
    W_{i,n}(\btheta)\,p_{i,0}(\mathrm d\btheta).
\end{align}
These quantities are used only in the analysis and need 
not be computed by the agent. Recall also that
\(\pi_n:=\pi^2n^2/3\), so that
\(\sum_{n\geqslant1}\pi_n^{-1}=1/2\).

\begin{proposition}\label{prop:fgts_bound}
    Suppose that Assumptions \ref{ass:FGTS-N'}, 
    \ref{ass:FGTS-P'} and the first part of 
    \Cref{ass:FGTS-F'} hold. 
    Fix \(i\in[K]\) and \(\delta\in(0,1/2)\). 
    With probability at least
    \(1-\delta\), the following bound holds simultaneously
    for all \(n\geqslant1\):
    \begin{align}\label{eq:fgts_regret_bound}
        \mathcal R_{i,n}
        &\leqslant \frac{\dc_i(\Theta_i)^2\bar\kappa_n}{\eta_i} \lambda_{i,n} n
        -\frac{\widehat Z_{i,n}}{\lambda_{i,n}} 
        +\frac{\log(2/\delta)}{\lambda_{i,n}}
        +\mathsf A_i
        \sqrt{\frac{n}2\log\frac{\pi_n}{\delta}}.
    \end{align}
\end{proposition}
The log-partition term relates regret to the prior mass
assigned to accurate and sufficiently optimistic candidates.
The remaining two terms arise from concentration: the first
controls the posterior mixture uniformly over time, while the
second converts conditional expected regret into realized
regret.

\begin{proof}[Proof of \Cref{prop:fgts_bound}]
Since \(i\in[K]\) is fixed, we suppress this index throughout
the proof. The argument has four steps. We first
separate the conditional regret into feel-good and prediction
error terms. We then relate the posterior mixture to the
log-partition function, bound its conditional moment, and
finally pass to realized regret.

\paragraph{Conditional information}
On the local clock of the fixed agent, write
\begin{align}\label{eq:fgts_bound_proof_filtrations}
    \mathcal F_{j-1}:=\sigma(H_{\tau_j-1},\mathbf C_j),
    \qquad \mathcal F_{j-1}^{x}
    :=\sigma(H_{\tau_j-1},\mathbf C_j,\mathbf X_j).
\end{align}
The first \(\sigma\)-algebra contains the information available 
before sampling the action; the second also contains the 
selected action. The next \(\mathcal F_j\) 
contains the reward from visit \(j\), and the posterior 
\(p_j\) is \(\mathcal F_{j-1}\)-measurable. We work 
on the local clock of the fixed agent under the stated 
conditional sampling and noise assumptions. The
bound holds simultaneously over visit counts and can
therefore be evaluated at the random count \(N_i(T)\)
in the meta-level application.

We introduce the quantities
\begin{align}
    \mathrm{REG}_j:=f^\star(\mathbf C_j)
    -f(\mathbf X_j,\mathbf C_j),\qquad 
    D_j^R:=\mathrm{REG}_j
-\mathbb E[\mathrm{REG}_j\mid\mathcal F_{j-1}]
\end{align}
that allow us to decompose the cumulative regret as
follows:
\begin{align}
    \mathcal R_n =\sum_{j=1}^n \mathbb E[\mathrm{REG}_j
    \mid\mathcal F_{j-1}] + \sum_{j=1}^n D_j^R. 
\end{align}
\paragraph{Step 1: decoupling the conditional regret}
For posterior integration, introduce \(\tilde\bvartheta_j\)
with conditional law \(p_j\), using fresh randomness independent
of \((\bvartheta_j,R_j)\) given \(\mathcal F_{j-1}\).
Thus the two parameter draws are conditionally independent,
and the auxiliary draw is also independent of the reward
conditionally on \(\mathcal F_{j-1}^{x}\). It is used only
in the analysis.
Define the feel-good and the Bellman errors:
\begin{align}\label{eq:fgts_regr3}
    \mathrm{FG}_j(\btheta)
    &:=f^\star(\mathbf C_j)-f_{\btheta}(\mathbf C_j),\quad
    \text{and}\quad 
    \mathrm{BE}_j(\btheta,\tilde\btheta)
    &:=f_{\tilde\btheta}\big(\mathbf x(\btheta,\mathbf C_j),
    \mathbf C_j\big)
    -f\big(\mathbf x(\btheta,\mathbf C_j),\mathbf C_j\big).
\end{align}
Since \(\mathbf X_j=\mathbf x(\bvartheta_j,\mathbf C_j)\),
the per-visit regret satisfies
\begin{align}\label{eq:fgts_regr2}
    \mathrm{REG}_j:=f^\star(\mathbf C_j)
    -f(\mathbf X_j,\mathbf C_j)
    =\mathrm{FG}_j(\bvartheta_j)
    +\mathrm{BE}_j(\bvartheta_j,\bvartheta_j).
\end{align}
Set
\begin{align}
    \overline{\mathrm{FG}}_j
    &:=\mathbb E\big[\mathrm{FG}_j(\tilde\bvartheta_j)
    \bigm|\mathcal F_{j-1}\big],\qquad\text{and}\qquad
    \overline{\mathrm{LS}}_j
    :=\big(\mathbb E\big[
    \mathrm{BE}_j(\bvartheta_j,\tilde\bvartheta_j)^2
    \bigm|\mathcal F_{j-1}\big]\big)^{1/2}.
    \label{eq:fgts_regr4}
\end{align}
Both quantities are \(\mathcal F_{j-1}\)-measurable,
and \(0\leqslant\overline{\mathrm{LS}}_j\leqslant\mathsf A\).
The actual and auxiliary parameter draws, \(\bvartheta_j\) 
and \(\tilde\bvartheta_j\), have the same
conditional law, so they give the same posterior mean of
\(\mathrm{FG}_j\).  This leads to
\begin{align}
    \mathcal R_n &= \sum_{j=1}^n\big( \overline{\mathrm{FG}}_j 
    + \mathbb E\big[\,\mathrm{BE}_j(\bvartheta_j, \bvartheta_j)
    \mid \mathcal F_{j-1}\big]\big) + \sum_{j=1}^n D_j^R.
\end{align}
The expression in parentheses equals \(\mathbb E[ 
\mathrm{REG}_j\mid\mathcal F_{j-1}]\) and is therefore nonnegative. Since \(\lambda_j\geqslant\lambda_n\) for 
\(j\leqslant n\), we obtain
\begin{align}
    \mathcal R_n 
    &\leqslant \frac1{\lambda_n}\sum_{j=1}^n 
    \lambda_j\big(\overline{\mathrm{FG}}_j +
    \mathbb E\big[\,\mathrm{BE}_j(\bvartheta_j,
    \bvartheta_j)\mid \mathcal F_{j-1}\big]\big) + 
    \sum_{j=1}^n D_j^R.
\end{align}
Applying \eqref{eq:fgts_dc_def} to \(q=p_j\) yields \(\mathbb 
E[\,\mathrm{BE}_j(\bvartheta_j, \bvartheta_j) \mid\mathcal 
F_{j-1}] \leqslant \dc(\Theta)\, \overline{\mathrm{LS}}_j\). 
Therefore, 
\begin{align}\label{eq:fgts_1}
    \mathcal R_n \leqslant \frac1{\lambda_n}\sum_{j=1}^n 
    \lambda_j \overline{\mathrm{FG}}_j +
    \frac1{\lambda_n}\sum_{j=1}^n \lambda_j
    \dc(\Theta)\,\overline{\mathrm{LS}}_j + 
    \sum_{j=1}^n D_j^R.
\end{align}
It remains to control the posterior mean feel-good term while
retaining a negative squared-error term that offsets the
second term on the right. This is done using the following
decomposition. Write \(\Delta L_j(\btheta)
:=\Delta L_j(\btheta,\mathbf C_j,\mathbf X_j,R_j)\), and define
\begin{align}\label{eq:fgts_mixture_definitions}
    G_j:=\int_\Theta e^{-\Delta L_j(\btheta)}
    p_j(\mathrm d\btheta),
    \qquad \overline G_j:=\mathbb E[G_j\mid\mathcal F_{j-1}].
\end{align}
The moment bound in Step 3 establishes that
\(0<\overline G_j<\infty\) almost surely. Also,
\(0<G_j<\infty\) pathwise because predictions are bounded
and the observed reward is finite. We can now rewrite 
\eqref{eq:fgts_1} in the following form:
\begin{align}\label{eq:fgts_regr5}
    \mathcal R_n \leqslant \frac1{\lambda_n}\sum_{j=1}^n 
    \big(\lambda_j \overline{\mathrm{FG}}_j + \log 
    \overline{G}_j\big) +  \frac1{\lambda_n}
    \sum_{j=1}^n \lambda_j \dc(\Theta)\,\overline{\mathrm{LS}}_j 
    + \sum_{j=1}^n D_j^R - \frac1{\lambda_n}\sum_{j=1}^n
    \log \overline{G}_j.
\end{align}

\paragraph{Step 2: the posterior mixture and the partition 
function}
The goal of this step is to upper bound the sum over \(j\)
of the terms \(-\log\overline{G}_j\), the last terms
appearing in decomposition \eqref{eq:fgts_regr5}. 
Set \(W_0(\btheta):=1\) and \(\widehat Z_0:=0\). The
terms subtracted from \(L_j\) to form \(\Delta L_j\) do
not depend on \(\btheta\), and hence cancel in the posterior
normalization. Consequently,
\begin{align}\label{eq:fgts_posterior_weights}
    W_j(\btheta)=W_{j-1}(\btheta)e^{-\Delta L_j(\btheta)},
    \qquad p_j(\mathrm d\btheta)
    =e^{-\widehat Z_{j-1}}W_{j-1}(\btheta)
    p_0(\mathrm d\btheta).
\end{align}
Integrating the weight recursion gives
\begin{align}\label{eq:fgts_partition_recursion}
    e^{\widehat Z_j}
    &=\int_\Theta W_{j-1}(\btheta)e^{-\Delta L_j(\btheta)}
    p_0(\mathrm d\btheta) =e^{\widehat Z_{j-1}}\int_\Theta
    e^{-\Delta L_j(\btheta)}p_j(\mathrm d\btheta)
    =e^{\widehat Z_{j-1}}G_j.
\end{align}
The second equality substitutes the posterior expression from
\eqref{eq:fgts_posterior_weights}. Taking logarithms and summing,
we obtain the exact identity
\begin{align}\label{eq:fgts_regr7}
    \widehat Z_n=\sum_{j=1}^n\log G_j.
\end{align}
To compare the mixture with its conditional mean, define
\(M_0:=1\) and \(M_n:=M_{n-1}G_n/\overline G_n\). Since
\(M_{n-1}\) and \(\overline G_n\) are
\(\mathcal F_{n-1}\)-measurable,
\begin{align}
    \mathbb E[\,M_n\mid\mathcal F_{n-1}]
    =\frac{M_{n-1}}{\overline G_n}\,
    \mathbb E[\,G_n\mid\mathcal F_{n-1}]=M_{n-1}.
\end{align}
Thus \((M_n)\) is a nonnegative martingale starting at one.
Ville's inequality~\citep{ville1939etude,howard2020time} gives an event of
probability at least \(1-\delta/2\) on which
\(M_n\leqslant2/\delta\) for every \(n\geqslant1\).
Using \eqref{eq:fgts_regr7}, on this event we have
\begin{align}\label{eq:fgts_predictable_partition_bound}
    -\sum_{j=1}^n\log\overline G_j = -\widehat Z_n +\sum_{j=1}^n
    \log (G_j/\overline{G}_j) = -\widehat Z_n + \log M_n
    \leqslant -\widehat Z_n +\log(2/\delta),
    \quad n\geqslant1.\quad 
\end{align}
This is the only concentration step needed for the
log-partition function itself. The upper bound
\eqref{eq:fgts_regr5} can be further bounded by
\begin{align}\label{eq:fgts_2}
    \mathcal R_n \leqslant \frac1{\lambda_n}\sum_{j=1}^n 
    \big(\lambda_j \overline{\mathrm{FG}}_j + \log 
    \overline{G}_j\big) + \frac1{\lambda_n}\sum_{j=1}^n 
    \lambda_j\dc(\Theta)\,\overline{\mathrm{LS}}_j + 
    \sum_{j=1}^n D_j^R - \frac{\widehat Z_n}{\lambda_n}
    +\frac{\log(2/\delta)}{\lambda_n}.
\end{align}

\paragraph{Step 3: controlling the conditional mixture}
The following lemma supplies the negative squared-error term
needed in \eqref{eq:fgts_regr5}.

\begin{lemma}\label{lem:fgts_mixture}
    Assume \eqref{eq:fgts_noise_gen} and 
    \eqref{eq:fgts_bounded_rewards_gen}, with the index 
    suppressed, and let \(\eta>0\) satisfy
    \(4\eta(\sigma^2\vee\mathsf A^2)\leqslant 1\).
    Then \(0<\overline G_j<\infty\) almost surely and
    \begin{align}\label{eq:fgts_mixture_bound}
        \lambda_j \overline{\mathrm{FG}}_j
        + \log\overline G_j
        \leqslant\frac{\mathsf A^2\lambda_j^2}{4}
        -\frac{5\eta}{16}\,\overline{\mathrm{LS}}_j^2.
    \end{align}
\end{lemma}
We postpone the proof of this lemma to the end of this 
section.  Combining the lemma with \eqref{eq:fgts_2} and 
completing the square gives
\begin{align}
    \mathcal R_n &\leqslant \frac1{\lambda_n}\sum_{j=1}^n 
    \Big(\frac{\mathsf A^2\lambda_j^2}{4} - \frac{5\eta}{16}
    \overline{\mathrm{LS}}_j^2 + \lambda_j\dc(\Theta)\,\overline{\mathrm{LS}}_j\Big) +
    \sum_{j=1}^n D_j^R - \frac{\widehat Z_n}{\lambda_n}
    +\frac{\log(2/\delta)}{\lambda_n}\\
    &\leqslant \frac{\mathsf A^2\bar\kappa_n}{4} \, \lambda_n n
    + \frac{0.8\bar\kappa_n}{\eta} \dc(\Theta)^2\lambda_n n
     + 
    \sum_{j=1}^n D_j^R - \frac{\widehat Z_n}{\lambda_n}
    +\frac{\log(2/\delta)}{\lambda_n}
    \label{eq:fgts_3}
\end{align}
Indeed, for every \(u,d\geqslant0\),
\begin{align}
    du-\frac{5\eta}{16\lambda}u^2
    =-\frac{5\eta}{16\lambda}
    \Big(u-\frac{8\lambda d}{5\eta}\Big)^2
    +\frac{4\lambda d^2}{5\eta}
    \leqslant\frac{0.8\lambda}{\eta}d^2.
\end{align}
By definition, \(\dc(\Theta)\geqslant1\). Since
\(4\eta\mathsf A^2\leqslant1\), we have
\begin{align}
    \frac{\mathsf A^2}{4}
    +\frac{4\dc(\Theta)^2}{5\eta}
    \leqslant
    \frac{69}{80\eta}\dc(\Theta)^2
    \leqslant\frac{\dc(\Theta)^2}{\eta}.
\end{align}
This leads to
\begin{align}
    \mathcal R_n 
    &\leqslant \frac{\dc(\Theta)^2\bar\kappa_n}{\eta} \, \lambda_n n
     + 
    \sum_{j=1}^n D_j^R - \frac{\widehat Z_n}{\lambda_n}
    +\frac{\log(2/\delta)}{\lambda_n}
    \label{eq:fgts_4}
\end{align}
The square completion explains the role of decoupling: its
linear prediction-error term is absorbed by the negative
quadratic term supplied by the posterior mixture.

\paragraph{Step 4: passing to realized regret}
Since \(0\leqslant\mathrm{REG}_j\leqslant\mathsf A\),
the variables \(D_j^R\) are martingale differences whose
conditional ranges have length \(\mathsf A\). The 
Azuma-Hoeffding inequality implies that for each 
fixed \(n\geqslant1\),
\begin{align}
    \mathbb P\bigg(\sum_{j=1}^n 
    D_j^R> \mathsf A\sqrt{\frac{n}2\log 
    \frac{\pi_n}{\delta}}\bigg)\leqslant\frac\delta{\pi_n}.
\end{align}
A union bound and \(\sum_{n\geqslant1}\pi_n^{-1}=1/2\)
show that these deviations are bounded simultaneously for
all \(n\) with probability at least \(1-\delta/2\). 
Combining with \eqref{eq:fgts_4}, we get 
\begin{align}
    \mathcal R_n 
    &\leqslant \frac{\dc(\Theta)^2\bar\kappa_n}{\eta} \, \lambda_n n
     - \frac{\widehat Z_n}{\lambda_n}
    +\frac{\log(2/\delta)}{\lambda_n} + \mathsf A\sqrt{\frac{n}2
    \log \frac{\pi_n}{\delta}}
    \label{eq:fgts_5}
\end{align}
with probability at least \(1-\delta\). This proves
\eqref{eq:fgts_regret_bound}. 
\end{proof}
\begin{proof}[Proof of \Cref{lem:fgts_mixture}]
    For fixed \(\btheta\), write
    \(E_j(\btheta):=f_{\btheta}(\mathbf X_j,\mathbf C_j)
    -f(\mathbf X_j,\mathbf C_j)\). Expanding the squared-loss
    difference gives
    \begin{align}\label{eq:fgts_loss_expansion}
        \Delta L_j(\btheta)
        =\eta E_j(\btheta)^2-2\eta E_j(\btheta)\varepsilon_j
        +\lambda_j\mathrm{FG}_j(\btheta).
    \end{align}
    Both \(E_j(\btheta)\) and \(\mathrm{FG}_j(\btheta)\)
    are \(\mathcal F_{j-1}^{x}\)-measurable. Applying the
    conditional sub-Gaussian bound at
    \(\rho=2\eta E_j(\btheta)\) therefore yields
    \begin{align}\label{eq:fgts_bound_proof_noise_moment}
        \mathbb E\big[e^{-\Delta L_j(\btheta)}
        \bigm|\mathcal F_{j-1}^{x}\big]
        \leqslant\exp\big(-\lambda_j \mathrm{FG}_j 
        (\btheta) -\eta(1-2\eta \sigma^2) E_j(\btheta)^2 
        \big).
    \end{align}
    The tuning condition implies \(4\eta\sigma^2\leqslant 1\), 
    so \(\eta(1-2\eta\sigma^2)\geqslant\eta/2\).
    Integrating over the auxiliary draw and then over the
    action, using conditional independence and the tower
    property, gives
    \begin{align}\label{eq:fgts_mixture_joint_moment}
        \overline G_j\leqslant\mathbb E\big[
        e^{-\lambda_j\mathrm{FG}_j(\tilde\bvartheta_j)
        -(\eta/2)\mathrm{BE}_j(\bvartheta_j, 
        \tilde\bvartheta_j)^2} \bigm| \mathcal F_{j-1} 
        \big].
    \end{align}
    Conditional Tonelli justifies interchanging the posterior
    integral and conditional expectation. The right-hand side
    is finite since both terms in its exponent are bounded.
    Strict positivity follows from positivity of the integrand
    defining \(G_j\).

    Conditional Cauchy--Schwarz now gives
    \begin{align}\label{eq:fgts_mixture_cs}
        \log\overline G_j
        &\leqslant\frac12\log\mathbb E\big[
        e^{-2\lambda_j\mathrm{FG}_j(\tilde\bvartheta_j)}
        \bigm|\mathcal F_{j-1}\big]+\frac12\log\mathbb E\big[
        e^{-\eta\mathrm{BE}_j(\bvartheta_j,\tilde\bvartheta_j)^2}
        \bigm|\mathcal F_{j-1}\big].
    \end{align}
    Conditionally on \(\mathcal F_{j-1}\), the feel-good term
    lies in \([f^\star(\mathbf C_j)-\mathsf A,
    f^\star(\mathbf C_j)]\), an interval of length
    \(\mathsf A\). Conditional Hoeffding's lemma implies
    \begin{align}\label{eq:fgts_fg_hoeffding}
        \frac12\log\mathbb E\big[
        e^{-2\lambda_j\mathrm{FG}_j(\tilde\bvartheta_j)}
        \bigm|\mathcal F_{j-1}\big]
        \leqslant-\lambda_j\overline{\mathrm{FG}}_j
        +\frac{\mathsf A^2\lambda_j^2}{4}.
    \end{align}
    For the second term, bounded predictions imply
    \(0\leqslant\mathrm{BE}_j^2\leqslant\mathsf A^2\).
    Since \(\eta\mathsf A^2<1\), we may use the endpoint
    chord of \(u\mapsto e^{-u}\) on \([0,1]\). It gives
    \begin{align}\label{eq:fgts_exp_chord}
        e^{-\eta y}\leqslant1-(1-e^{-1})\eta y
        \leqslant1-\frac{5\eta}{8}y,
        \qquad 0\leqslant y\leqslant\mathsf A^2.
    \end{align}
    The second inequality uses \(1-e^{-1}>5/8\).
    Taking conditional expectations and applying
    \(\log(1-u)\leqslant-u\), we obtain
    \begin{align}\label{eq:fgts_ls_chord}
        \frac12\log\mathbb E\big[
        e^{-\eta\mathrm{BE}_j(\bvartheta_j,\tilde\bvartheta_j)^2}
        \bigm|\mathcal F_{j-1}\big]
        \leqslant\frac12\log\Big(1-\frac{5\eta}{8}\,
        \overline{\mathrm{LS}}_j^2\Big)
        \leqslant-\frac{5\eta}{16}\,\overline{\mathrm{LS}}_j^2.
    \end{align}
    The logarithm is well-defined because
    \((5\eta/8)\,\overline{\mathrm{LS}}_j^2
    \leqslant5\eta\mathsf A^2/8<1\).
    Substituting \eqref{eq:fgts_fg_hoeffding} and
    \eqref{eq:fgts_ls_chord} into \eqref{eq:fgts_mixture_cs}
    proves the claim.
\end{proof}

\subsection{High-probability regret bounds under 
misspecification}

Recall that \(\pi_n:=\pi^2n^2/3\), so that 
\(\sum_{n\geqslant1}\pi_n^{-1}=1/2\).
\begin{proposition}\label{prop:fgts_newbound}
    Suppose that Assumptions \ref{ass:FGTS-N'}, 
    \ref{ass:FGTS-F'} and \ref{ass:FGTS-P'} hold. Fix \(i\in 
    [K]\), \(\delta\in (0,1/2)\), and a deterministic 
    sequence of probability measures \((\bar p_{i,n})_{n
    \geqslant1}\) on \(\Theta_i\), with \(\bar p_{i,n}\ll 
    p_{i,0}\) for every \(n\geqslant1\). Set
    \begin{align}
        \bar f_{i,p}
        :=\int_{\Theta_i} f_{i,\btheta}\,p_i(\mathrm 
        d\btheta),
        \quad
        \zeta_{1,i}(p_i):=\|f_i-\bar f_{i,p}\|_\infty,
        \quad
        \zeta_{2,i}(p_i)
        :=\Big\|\int_{\Theta_i}(f_i-f_{i,\btheta})^2
        p_i(\mathrm d\btheta)\Big\|_\infty^{1/2}.
    \end{align}  
    With probability 
    at least \(1-\delta\), the following bound holds
    simultaneously for all \(n\geqslant1\):
    \begin{align}
        \mathcal R_{i,n}
        &\leqslant \frac{m_i\bar\kappa_n}{\eta_i}\lambda_{i,
        n} n +\frac{\eta_in}{\lambda_{i,n}}\zeta_{2,i}^2(\bar p_{i,n})
        +\frac{n\bar\lambda_{i,n}}{\lambda_{i,n}}\,\zeta_{1,
        i}(\bar p_{i,n}) 
        +\Big(\mathsf A_i + \frac{4\eta_i\sigma_i}{\lambda_{i,n}}\zeta_{1,i}(\bar p_{i,n})\Big)
        \sqrt{\frac{n}2\log\frac{2\pi_n}{\delta}}\\
        &\qquad+\frac{\log(2/\delta) + D_{\mathrm{KL}}(\bar p_{i,n} 
        \Vert p_{i,0})}{\lambda_{i,n}}.\label{eq:fgts_reg_b}
    \end{align}
\end{proposition}
The proof of this proposition consists in combining  
\Cref{prop:fgts_bound} and an upper bound on the
log-partition function established below. Since the latter 
deals with a fixed agent \(i\in[K]\),  for simplicity of
exposition, we remove indices \(i\) from the notation. 

\begin{lemma}\label{lem:fgts_partition_comparator}
    Fix a sequence of deterministic probability measures 
    \(\bar p_n\ll p_0\) on \(\Theta\), and set
    \begin{align}
        \bar f_p
        :=\int_\Theta f_{\btheta}\,\bar p_n(\mathrm d\btheta),
        \qquad
        \zeta_1(\bar p_n):=\|f-\bar f_p\|_\infty,
        \qquad
        \zeta_2(\bar p_n)
        :=\Big\|\int_\Theta(f-f_{\btheta})^2
        \bar p_n(\mathrm d\btheta)\Big\|_\infty^{1/2}.
    \end{align}
    Suppose \Cref{ass:FGTS-N'} and the first part
    of \Cref{ass:FGTS-F'} hold. Then, for every \(\delta
    \in(0,1)\), with probability at least \(1-\delta/2\), 
    simultaneously for all \(n\geqslant1\),
    \begin{align}
        -\widehat Z_n &\leqslant \eta \zeta_2(\bar p_n)^2n
        +\zeta_1(\bar p_n)\sum_{j=1}^n\lambda_j
        +2\eta\sigma\zeta_1(\bar p_n)
        \sqrt{2n\log({\pi_n}/{\delta})} + 
        D_{\mathrm{KL}}(\bar p_n\Vert p_0).
    \end{align}
\end{lemma}

\begin{proof}
    Recall that
    \begin{align}
        \widehat Z_n &=  \log\int_{\Theta} 
        e^{-\sum_{j=1}^n\Delta L_j(\btheta)}
        p_0(\mathrm d\btheta) \geqslant -
        \sum_{j=1}^n 
        \int_{\Theta} \Delta L_j(\btheta)\, 
        \bar p_n(\mathrm d\btheta) - 
        D_{\mathrm KL}(\bar p_n\Vert p_0),
    \end{align}
    where the last inequality follows from the 
    Donsker-Varadhan variational formula. 
    Expanding the squared-loss difference 
    gives
    \begin{align}\label{eq:fgts_loss_exp}
        \Delta L_j(\btheta) =\eta \big(
        f_{\btheta}(\mathbf X_j,\mathbf C_j) 
        -f(\mathbf X_j,\mathbf C_j)\big)^2 
        &+ \lambda_j
        \big(f^\star(\mathbf C_j)-f_{\btheta}
        (\mathbf C_j)\big)\\
        &- 2\eta \big(f_{\btheta}(\mathbf X_j,
        \mathbf C_j) - f(\mathbf X_j,\mathbf 
        C_j)\big)\varepsilon_j .
    \end{align}
    Integrating this equality and using the 
    notation \(\bar f_p \), \(\zeta_1(p)\) and 
    \(\zeta_2(p)\) introduced in the statement 
    of the lemma, we get
    \begin{align}\label{eq:fgts_loss_exp2}
        \int_\Theta \Delta L_j(\btheta)\,\bar p_n
        (\mathrm d\btheta)\leqslant \eta 
        \zeta_2^2(\bar p_n) + \lambda_j\zeta_1(\bar p_n)
        - 2\eta \big(\bar f_p(\mathbf X_j,
        \mathbf C_j) - f(\mathbf X_j,\mathbf 
        C_j)\big)\varepsilon_j .
    \end{align}
    Furthermore, the noise assumption and the 
    tower property give
    \begin{align}\label{eq:fgts_fin_mgf}
        \mathbb E\Big[\exp\Big(-\rho\sum_{j=1}^n
        \big(\bar f_p(\mathbf X_j,
        \mathbf C_j) - f(\mathbf X_j,\mathbf 
        C_j)\big)\varepsilon_j\Big)\Big]
        \leqslant e^{\rho^2\sigma^2\zeta_1(\bar p_n)^2n/2},
        \qquad \rho\in\mathbb R.
    \end{align}
    Chernoff's bound implies
    \begin{align}\label{eq:fgts_fin_cross_hp}
        \mathbb P\bigg(-\sum_{j=1}^n
        \big(\bar f_p(\mathbf X_j, \mathbf C_j) - 
        f(\mathbf X_j,\mathbf C_j)\big)\varepsilon_j
        >\sigma\zeta_1(\bar p_n)\sqrt{2n\log({\pi_n}/{\delta}
        )}\bigg) \leqslant\frac\delta{\pi_n}.
    \end{align}
    This implies that
    \begin{align}
        -\widehat Z_n \leqslant n\eta \zeta_2^2(\bar p_n)
        +\zeta_1(\bar p_n)\sum_{j=1}^n\lambda_j +2\sigma
        \eta \zeta_1(\bar p_n) \sqrt{2n\log({\pi_n}/{\delta}
        )} + D_{\mathrm{KL}}(\bar p_n\Vert p_0)
    \end{align}
    and the claim of the lemma follows.
\end{proof}

\subsection{Proof of \Cref{th:fgts_discrete}}
\label{sec:fgts_discr}

Fix \(i\in[K]\) and suppress this index throughout the
proof. The assumptions of the main text imply the
general assumptions with \(\mathsf A=\sigma=1\).
Moreover, the linear representation of the prediction
errors gives \(\dc(\Theta)^2\leqslant m\).

Let \(\Theta^\star:=\{\btheta\in\Theta:
f_{\btheta}\equiv f\}\), so that
\(p_0(\Theta^\star)=q^\star>0\). For every
\(\btheta\in\Theta^\star\), both terms in the excess
loss vanish. Consequently,
\begin{align}
    \Delta L_j(\btheta)=0,
    \qquad W_n(\btheta)=1,
    \qquad \btheta\in\Theta^\star.
\end{align}
Restricting the partition integral to \(\Theta^\star\)
therefore gives the pathwise bound
\begin{align}
    e^{\widehat Z_n}
    =\int_\Theta W_n(\btheta)\,p_0(\mathrm d\btheta)
    \geqslant\int_{\Theta^\star}
    W_n(\btheta)\,p_0(\mathrm d\btheta)
    =q^\star.
\end{align}
Applying \Cref{prop:fgts_bound}, we obtain, with
probability at least \(1-\delta\), simultaneously
for all \(n\geqslant1\),
\begin{align}
    \mathcal R_n
    \leqslant\frac{m\bar\kappa_n}{\eta}\lambda_n n
    +\frac{\log(2/\delta)+\log(1/q^\star)}{\lambda_n}
    +\sqrt{\frac n2\log\frac{\pi_n}{\delta}}.
\end{align}
Since \(\delta\leqslant 1/2\) and \(\pi_n=\pi^2n^2/3\), 
we have \(\tfrac12\log\tfrac{\pi_n}{\delta} \leqslant
\log\tfrac{2n}{\delta}\).  This proves the stated bound.

\subsection{Proof of \Cref{th:fgts_cont}}
\label{sec:fgts_cont}

Fix \(i\in[K]\) and suppress this index throughout the
proof. Write \(d:=d_i^\theta\), and choose
\(\btheta^\star\in\Theta=\mathbb B_1(0)\) such that
\(f_{\btheta^\star}=f\). The assumptions of the theorem
imply Assumptions \ref{ass:FGTS-N'},
\ref{ass:FGTS-F'}, and \ref{ass:FGTS-P'} with
\(\mathsf A=\sigma=1\).

For each \(n\geqslant1\), define
\begin{align}
    r_n:=\min\{2,1/(\gamma n)\},
    \qquad
    B_n:=\Theta\cap\mathbb B_{r_n}(\btheta^\star),
    \qquad
    \bar p_n:=p_0(\,\cdot\mid B_n).
\end{align}
We will apply \Cref{prop:fgts_newbound} to the
deterministic sequence \((\bar p_n)_{n\geqslant1}\).
We first verify that these measures are well-defined
and bound their KL divergences.

Set \(\mathbf z_n:=(1-r_n/2)\btheta^\star\). For every
\(\btheta\in\mathbb B_{r_n/2}(\mathbf z_n)\), the
triangle inequality gives
\begin{align}
    \|\btheta\|_2
    &\leqslant r_n/2+(1-r_n/2)\|\btheta^\star\|_2
    \leqslant1,\\
    \|\btheta-\btheta^\star\|_2
    &\leqslant r_n/2+(r_n/2)\|\btheta^\star\|_2
    \leqslant r_n.
\end{align}
Thus \(\mathbb B_{r_n/2}(\mathbf z_n)\subseteq B_n\).
Since \(p_0\) is uniform on the unit ball, comparison
of volumes yields \(p_0(B_n)\geqslant(r_n/2)^d>0\). 
Consequently, \(\bar p_n\) is well-defined,
\(\bar p_n\ll p_0\), and
\begin{align}
    D_{\mathrm{KL}}(\bar p_n\Vert p_0)
    =-\log p_0(B_n)
    \leqslant d\log(2/r_n)
    =d\log_+(2\gamma n).
\end{align}
Next, the Lipschitz assumption and realizability imply
\begin{align}
    \|f_{\btheta}-f\|_\infty
    \leqslant\gamma\|\btheta-\btheta^\star\|_2
    \leqslant\gamma r_n\leqslant1/n,
    \qquad \btheta\in B_n.
\end{align}
Since \(\bar p_n\) is supported on \(B_n\), the
definitions of the approximation errors give
\begin{align}
    \zeta_1(\bar p_n)\leqslant1/n,
    \qquad
    \zeta_2(\bar p_n)\leqslant1/n.
\end{align}
Substituting these inequalities and the KL bound into
\Cref{prop:fgts_newbound}, with \(\mathsf A=\sigma=1\),
shows that, with probability at least \(1-\delta\),
simultaneously for all \(n\geqslant1\),
\begin{align}
    \mathcal R_n
    &\leqslant\frac{m\bar\kappa_n}{\eta}\lambda_n n
    +\frac{\log(2/\delta)+d\log_+(2\gamma n)
    +\eta/n}{\lambda_n}
    +\frac{\bar\lambda_n}{\lambda_n}
    +\Big(1+\frac{4\eta}{n\lambda_n}\Big)
    \sqrt{\frac n2\log\frac{2\pi_n}{\delta}}.
\end{align}
Finally, \(\eta\leqslant1/4\) and \(n\geqslant1\)
imply
\begin{align}
    \log(2/\delta)+\eta/n
    \leqslant\log(2/\delta)+1/4
    \leqslant\log(3/\delta).
\end{align}
Furthermore, \(\pi_n=\pi^2n^2/3\) and
\(\delta<1/2\) give
\begin{align}
    \frac12\log\frac{2\pi_n}{\delta}
    =\log\frac{2n}{\delta}
    +\frac12\log\frac{\pi^2\delta}{6}
    \leqslant\log\frac{2n}{\delta}.
\end{align}
Combining these inequalities proves the theorem.

\section{Proofs of the Meta-Frieren bounds with FGTS}\label{sec:frieren_fgts}

We first verify \Cref{assump:regret_terms}, then use one
common calculation for the two regret envelopes. 

\paragraph{Noise and context concentration}
Fix \(i\) and \(n\). Iterating the conditional noise bound
in \eqref{eq:fgts_noise} gives
\begin{align}
    \mathbb E\Big[\exp\Big(\rho\sum_{j=1}^n
    \varepsilon_{i,j}\Big)\Big]
    \leqslant e^{\rho^2\sigma^2 n/2},
    \qquad \rho\in\mathbb R.
\end{align}
The two-sided sub-Gaussian tail therefore implies
\begin{align}
    \mathbb P\Big(\Big|\frac1n\sum_{j=1}^n
    \varepsilon_{i,j}\Big|
    >\sigma\sqrt{\frac{2\log(\pi_n/\delta)}n}\Big)
    \leqslant\frac{2\delta}{\pi_n}.
\end{align}
When \(\sigma=0\), the noise vanishes almost surely and the
bound is immediate. A union bound over \(n\geqslant1\)
proves~\eqref{eq:RwN} in \Cref{assump:regret_terms} with the radius
in \eqref{eq:fgts_meta_rad} and failure probability \(\delta\).

For contexts, the stated conditional sampling assumption gives
conditional mean \(\mu_i\) for
\(f_i^\star(\mathbf C_{i,j})\), and this variable lies in
\([0,\mathsf A]\). Conditional Hoeffding's lemma, followed by
the same union bound, yields
\begin{align}\label{eq:fgts_context_concentration}
    \mathbb P\Big(\Big|\frac1n\sum_{j=1}^n
    f_i^\star(\mathbf C_{i,j})-\mu_i\Big|
    >\mathsf A\sqrt{\frac{\log(\pi_n/\delta)}{2n}}\Big)
    \leqslant\frac{2\delta}{\pi_n}.
\end{align}
Thus,~\eqref{eq:OCR} holds with the
stated radius and failure probability \(\delta\). Independence
of contexts across indices is not needed for this calculation.
Finally, the tuned guarantees verify~\eqref{eq:ALL}. Each envelope is
deterministic and nondecreasing. Intersecting these three
events over the \(K\) indices gives the probability statement
in the main text.

\subsection{Proof of \Cref{th:frieren_fgts_finite}}\label{sec:frieren_fgts_finite_proof}

Work on \(\mathcal E(\delta)\), and fix \(T\geqslant K\).
For brevity, write \(u:=u(T)\), \(q:=u(T)+w(T)\), and
\(n_i:=n_i(T)\) as in~\eqref{eq:upper_bound_action}.
Monotonicity of the logarithmic factors gives, 
for every \(1\leqslant n\leqslant T\),
\begin{align}\label{eq:fgts_meta_envelope_bounds}
    B_i(n,\delta)\leqslant u\sqrt n,
    \qquad \frac{B_i(n,\delta)}n+2\mathrm{rad}_i(n,\delta)
    \leqslant\frac q{\sqrt n}.
\end{align}
For a suboptimal index, let
\(r_i:=\lfloor q^2/\Delta_i^2\rfloor+1\).
If \(r_i\leqslant T\), then \(q/\sqrt{r_i}<\Delta_i\),
so the defining criterion for \(n_i\) holds at \(r_i\).
If \(r_i>T\), the convention \(n_i\leqslant T\) suffices.
In either case,
\begin{align}\label{eq:fgts_meta_count_bound}
    n_i\leqslant1+\frac{q^2}{\Delta_i^2},
    \qquad \sqrt{n_i}\leqslant1+\frac q{\Delta_i}.
\end{align}
The additive one accounts for the strict inequality and the
integer visit count in \eqref{eq:upper_bound_action}.
Consequently,
\begin{align}\label{eq:fgts_finite_p1}
    \sum_{i\notin\mathcal I^\star}n_i\Delta_i
    \leqslant\|\boldsymbol\Delta\|_1
    +q^2\sum_{i\notin\mathcal I^\star}\frac1{\Delta_i},\\
    \sum_{i\notin\mathcal I^\star}B_i(n_i,\delta)
    \leqslant(K-|\mathcal I^\star|)u
    +uq\sum_{i\notin\mathcal I^\star}\frac1{\Delta_i}.
    \label{eq:fgts_finite_p2}
\end{align}
For optimal indices, any nonnegative integer allocation with
\(\sum_{i\in\mathcal I^\star}u_i\leqslant T\) satisfies
\begin{align}\label{eq:fgts_finite_p3}
    \sum_{i\in\mathcal I^\star}B_i(u_i,\delta)
    \leqslant u\sum_{i\in\mathcal I^\star}\sqrt{u_i}
    \leqslant u\sqrt{T|\mathcal I^\star|}.
\end{align}
The last inequality is Cauchy--Schwarz. Substituting these
three bounds into \eqref{eq:regret_meta_bound} proves
\eqref{eq:fgts_finite_meta_regret}.

\subsection{Proof of \Cref{th:frieren_fgts_parametric}}\label{sec:frieren_fgts_parametric_proof}

The continuous-class envelope satisfies~\eqref{eq:fgts_meta_envelope_bounds} with
\(u=u^\theta(T)\) and \(q=u^\theta(T)+w(T)\).
The noise and context radii are unchanged. Applying~\eqref{eq:fgts_meta_count_bound} and the three subsequent
bounds with these substitutions proves~\eqref{eq:fgts_parametric_meta_regret}.


\section{Regret Conversion}\label{sec:reg_conv}
We define 
\begin{align}
    \Lambda(\boldsymbol{f}):=\max_{i\in[K]} 
    \sup_{\mathbf{c}\in\mathfrak{C}_i}\{\sup_{\mathbf{x}\in\mathfrak{X}_i} f_i(\mathbf{x}, \mathbf{c})-
    \inf_{\mathbf{x}\in\mathfrak{X}_i}f_i(\mathbf{x}, \mathbf{c})\}.
\end{align}
The following proposition relates \(\mathcal R(T)\)
and \(\widehat{\mathcal R}(T)\).
\begin{proposition}\label{prop:reg_conv}
    Fix \(\eta\in(0,1)\). Then, 
    \begin{align}\label{eq:reg_conv}
        \widehat{\mathcal{R}}(T)\leqslant 
        2\left(\mathcal{R}(T)+\Lambda(\boldsymbol{f})\log(1/\eta)\right),
    \end{align}
    with probability at least \(1-\eta\), simultaneously for all \(T\in \mathbb{N}^+\).
\end{proposition}
\begin{proof}
    Let \(\mathcal{F}_t\) be the \(\sigma\)-algebra generated by \((H_{t}, \boldsymbol{f})\) and
    \begin{align}
        \ell_t:=f_{I_t}^\star(\mathbf{C}_t)&-f_{I_t}(\mathbf{X}_t, \mathbf{C}_t), 
        \qquad 
        q_t:=\mathbb{E}[\ell_t\mid \mathcal{F}_{t-1}],\\
        A(T)&:=\sum_{t=1}^T\ell_t,
        \qquad 
        Q(T):=\sum_{t=1}^T q_t.
    \end{align}
    Recall that \(\mathcal R(T)\) and \(\widehat{\mathcal R}(T)\) differ only in
    their action selection components~\eqref{eq:our_regret},~\eqref{eq:weak_regret_open}, so that
    \begin{align}\label{app:eq:reg_diff}
        \widehat{\mathcal R}(T) - \mathcal R(T)=Q(T)-A(T).
    \end{align}
    Since \(\Lambda(\boldsymbol{f})\) is \(\mathcal{F}_0\)-measurable, 
    \(0\leqslant \ell_t, q_t\leqslant \Lambda(\boldsymbol{f})\). 
    We consider
    \begin{align}
        \mathbb{E}[\exp(-\ell_t/\Lambda(\boldsymbol{f}))\mid \mathcal{F}_{t-1}]&\leqslant 1-(1-e^{-1})q_t/\Lambda(\boldsymbol{f})\\
        &\leqslant \exp(-(1-e^{-1})q_t/\Lambda(\boldsymbol{f})),
    \end{align} 
    where we have used the endpoint chord of \(u\mapsto e^{-u}\) on \([0,1]\) with \(0/0:=0\) convention. 
   So
    \begin{align}
        M_T:=\exp\left(\frac{(1-e^{-1})Q(T)-A(T)}{\Lambda(\boldsymbol{f})}\right), \quad M_0=1,
    \end{align}
    is a nonnegative supermartingale. Ville's inequality gives,
    with probability at least \(1-\eta\), simultaneously for all
    \(T\geqslant1\),
    \[
        (1-e^{-1})Q(T)
        \leqslant A(T)+\Lambda(\boldsymbol f)\log(1/\eta).
    \]
    Using \(A(T)\leqslant\mathcal R(T)\),
    \(1-e^{-1}>1/2\), and~\eqref{app:eq:reg_diff}
    yields~\eqref{eq:reg_conv}.
        
\end{proof}

In the frequentist setting of \Cref{sec:freq_gp_ucb},
it follows from~\eqref{eq:freq_function_bound} that \(\Lambda(\boldsymbol{f})\leqslant 2b\).
In the Bayesian setting of \Cref{sec:by_gp_ucb},
it follows from~\eqref{eq:grad_op_norm_f} that
\begin{align}
\Lambda(\boldsymbol{f})\leqslant \max_{j\in[K]}d_j^xw_j^x\varphi
\sqrt{\log(6Kd_j\psi/\delta)},
\end{align}
on the event \(\mathcal E(\delta)\).
Hence, in both settings, we obtain an upper bound on \(\Lambda(\boldsymbol{f})\)
which can be plugged into \Cref{prop:reg_conv}.

\section{Experimental details}
This section provides additional information on the experiments described in \Cref{sec:experiments}.

\subsection{FGTS implementation details}\label{app:fgts_exp_det}
Exact sampling of \(\bvartheta_{i,j}\) from the FGTS posterior~\eqref{eq:fgts_posterior} is
intractable for the parametric class of \Cref{sec:fgts_param_fnc}. We therefore approximate
each draw with stochastic gradient Langevin dynamics (SGLD)~\citep{welling2011bayesian},
targeting the potential
\begin{align}
    U_{i,j}\left(\btheta\right)
    =\sum_{s=1}^{j-1}
    L_i\left(\btheta,\mathbf{C}_{i,s},\mathbf{X}_{i,s},R_{i,s}\right)
    -\log p_{i,0}\left(\btheta\right),
\end{align}
with \(L_i\) as in~\eqref{eq:fgts_neg_log_likelihood}. The squared-error term of \(L_i\) is
implemented as a Gaussian negative log-likelihood whose observation scale \(\sigma_i\) is
appended to the chain as an additional coordinate, which amounts to sampling
\(\eta_i=1/\left(2\sigma_i^2\right)\) jointly with \(\btheta\). The feel-good term requires
the greedy value \(f_{i,\btheta}\left(\mathbf{C}_{i,s}\right)\). For every distinct context
in the minibatch, we approximate the maximization in~\eqref{eq:fgts_greedy_action} by
the U-NSGA-III evolutionary optimizer~\citep{seada2016unified}. The reported runs use \(\lambda_i=0.01\). 
In place of the uniform prior of \Cref{th:fgts_cont}, we use a
nearly flat Gaussian prior with precision \(10^{-4}\). Our implementation builds on the Banditry codebase~\citep{banditry}.

We run the preconditioned variant of SGLD~\citep{li2016preconditioned}. Let \(g_t\) denote an
unbiased minibatch estimate of \(\nabla U_{i,j}\left(\btheta_t\right)\) at iteration \(t\) of
the chain. The sampler maintains an exponential moving average of the squared gradients and
the induced diagonal preconditioner,
\begin{align}
    v_t=\alpha v_{t-1}+\left(1-\alpha\right)g_t\odot g_t,
    \qquad
    G_t=\mathrm{diag}\!\left(\left(\sqrt{v_t}+\epsilon_0\right)^{-1}\right),
\end{align}
and iterates
\begin{align}\label{eq:psgld_update}
    \btheta_{t+1}
    =\btheta_t-\epsilon_t\,G_t g_t
    +\sqrt{2\epsilon_t\tau}\,G_t^{1/2}\xi_t,
    \qquad
    \xi_t\sim\mathcal{N}\left(\mathbf 0,\mathbf I\right),
\end{align}
with \(\tau=0.2\), \(\alpha=0.99\) and \(\epsilon_0=10^{-5}\).
During burn-in, the step size follows the polynomial
schedule of~\citep{welling2011bayesian},
\(\epsilon_t=a\left(b+t\right)^{-\gamma}\) with \(a=5\cdot10^{-3}\), \(b=1\), and
\(\gamma=0.55\); after burn-in it is frozen at
\(\max\left\{a\left(b+t_{\mathrm{burn}}\right)^{-\gamma},10^{-3}\right\}\), where
\(t_{\mathrm{burn}}\) is the number of burn-in iterations, so that the chain keeps mixing.

\subsection{GP-UCB implementation details}\label{app:gp_ucb_exp_det}
The within-index bound~\eqref{eq:b_gpucb_B} involves the maximum information gain
\(\gamma_{i,n}\), which is known only up to constant factors for the kernels we use, namely
the squared-exponential and Mat{\'e}rn kernels~\citep{srinivas2009gaussian,vakili2021information}.
Our implementation therefore replaces it with the empirical counterpart
\begin{equation}\label{eq:realised_ig}
    \widehat{\gamma}_{i,n}
    :=\frac{1}{2}\sum_{j=1}^{n}
    \log\left(1+\sigma^{-2}s^2_{i,j-1}\left(\mathbf{X}_{i,j},\mathbf{C}_{i,j}\right)\right),
\end{equation}
Lemma~5.4 of~\citet{krause2011contextual} bounds the regret of GP-UCB by the
information gain of the observed points, and the maximum information gain enters their
analysis only as an upper bound on this quantity. Hence, the bound~\eqref{eq:b_gpucb_B}
remains valid under this replacement in both the Bayesian and the frequentist
configurations.

The issue of unspecified constants also arises for \(\beta_i\). In the frequentist
configuration, we therefore set
\begin{equation}\label{eq:empirical_beta}
    \beta_i(n, \delta)=\left(b+4\sigma\sqrt{\widehat{\gamma}_{i,n}+1+\log(1/\delta)}\right)^2.
\end{equation}
This choice provides the same theoretical guarantees as the one
in \Cref{sec:freq_gp_ucb}~\citep{berkenkamp2019no}. In the Bayesian configuration, we follow
HEBO~\citep{Cowen-Rivers2022-HEBO} and the Banditry codebase~\citep{banditry} and set
\(\beta_i(n, \delta)=\left(2+\tfrac{d_i}{2}\right)\log n+\log\left(\pi^2/\delta\right)\).

On each play, we maximize the acquisition function over \(\mathfrak{X}_i\), with the context
fixed to its observed value, using the mixed-variable U-NSGA-III evolutionary
optimizer~\citep{seada2016unified,blank2020pymoo}, with a population of \(100\) evolved
for \(100\) generations and initialized at the best previously observed action.

\subsection{RKHS norms of the reward functions}\label{app:rkhs_norm}

Let \(\mathcal{H}_i\) denote the RKHS induced by the squared-exponential kernel
\begin{align}
    k_i(z,z')
    =
    \exp\left(-\frac{\|z-z'\|_2^2}{2\ell_i^2}\right),
    \qquad
    z, z' \in\mathbb{R}^2.
\end{align}
Writing \(a=(1,-1)^\top\), the reward function satisfies \(
    f_i(z)
    =
    (a^\top z)^2 k_i(z,0).
\)
Let \(D_a\) denote the directional derivative along \(a\). Differentiating the
kernel with respect to its second argument gives
\begin{align}
    h_i(z):=D_{a,z'}^2 k_i(z,z')\big|_{z'=0}
    =
    \left(
    \frac{(a^\top z)^2}{\ell_i^4}
    -
    \frac{\|a\|_2^2}{\ell_i^2}
    \right)k_i(z,0).
\end{align}
Since \(\|a\|_2^2=2\), we may therefore write
\(f_i=\ell_i^4 h_i+2\ell_i^2 k_i(\cdot,0).\)
Using the reproducing property for kernel derivatives,
\begin{align}
    \|h_i\|_{\mathcal H_i}^2=\frac{12}{\ell_i^4},
    \qquad
    \langle h_i,k_i(\cdot,0)\rangle_{\mathcal H_i}
    =-\frac2{\ell_i^2},
    \qquad
    \|k_i(\cdot,0)\|_{\mathcal H_i}^2=1.
\end{align}
Consequently, \(
\|f_i\|_{\mathcal H_i}^2
=
8\ell_i^4
\). For \(\ell_1=2^{-3/4}\) and
\(\ell_2=\sqrt{3/(2\sqrt{2})}\), this yields \(
\|f_1\|_{\mathcal H_1}=1\) and \(
\|f_2\|_{\mathcal H_2}=3.
\)

\subsection{Variable-type handling in the MAGIC experiment}
\label{app:magic_variable_types}

We consider three types of hyperparameters in the search spaces of
Table~\ref{tab:magic_hpo_spaces}, which are handled identically in the Bayesian GP-UCB
and FGTS configurations. A \emph{continuous} variable is optimized directly over its
real-valued interval, with scale-sensitive quantities parameterized by their base-\(10\)
logarithm. An \emph{integer} variable is represented as a real-valued coordinate in the
reward model and rounded when decoded, whereas U-NSGA-III~\citep{seada2016unified}
treats it directly as an integer variable, ensuring that only integer values are
proposed. A \emph{categorical} variable with \(V\) levels is represented by an integer
code in \(\{0,\dots,V-1\}\). Within the reward model, this code is mapped through a
learned embedding into \(\mathbb{R}^{1+\lfloor V/2\rfloor}\) following~\citet{banditry, Cowen-Rivers2022-HEBO}, with the embedding
parameters learned jointly with the remaining model parameters. U-NSGA-III instead
treats the variable as a choice among the \(V\) codes, without imposing an ordering or
continuous relaxation on the categories.

For example, the distance metric of the \(k\)-NN model is categorical, with levels
\{Manhattan, Euclidean, Chebyshev\}. It is encoded by an element of \(\{0,1,2\}\) and
mapped to a learned two-dimensional vector within the reward model. The maximum number
of leaf nodes for GBDT is an integer in \(\{8,\dots,128\}\). The reward model represents
it as a real-valued coordinate, with the resulting proposal rounded before training.
Finally, the GBDT learning rate is continuous and scale-sensitive, and is therefore
optimized through its base-\(10\) logarithm over \([-3,0]\).


\end{document}